\documentclass[a4paper,fleqn]{cas-dc}

\usepackage[numbers,sort&compress]{natbib}
\usepackage{makecell}
\usepackage{capt-of}
\usepackage{algorithm}
\usepackage{algpseudocode}
\usepackage{tabularx}
\usepackage{array}
\usepackage{float}
\usepackage{placeins}
\usepackage{subcaption}

\begin{document}

\shorttitle{Adaptive Deadline- and Energy-aware Multi-UAV Caching}

\title[mode=title]{Adaptive Deadline- and Energy-Aware Caching Algorithms in Dynamic Multi-UAV Edge Networks}

\author[1]{Jian Kim}
\affiliation[1]{
  organization={Department of Computer Science and Engineering, Chungnam National University},
  city={Daejeon},
  country={Korea}
}
\author[1]{Ki-Il Kim}\corref{cor1}
\cortext[cor1]{Corresponding author: Ki-Il Kim}
\tnotetext[1]{This work was supported by Korea Institute of Planning and Evaluation for Technology in Food, Agriculture, and Forestry(IPET) through the Agriculture and Food Convergence Technologies Program for Research Manpower development, funded by the Ministry of Agriculture, Food and Rural Affairs(MAFRA)(project no. RS-2024-00397026)}
\begin{abstract}
Content caching via unmanned aerial vehicles (UAVs) becomes one of the solution to maintain content delivery when terrestrial infrastructure is unavailable or disrupted. However, existing caching approaches often focus primarily on cache-hit performance or average delivery delay, without jointly addressing spatially heterogeneous and time-varying demand, content-specific deadlines, and uneven communication-energy consumption across UAVs. To address these challenges, we formulate a dynamic deadline- and energy-aware cache placement and replacement problem that jointly considers deadline misses, UAV-side communication energy, and inter-UAV energy imbalance. To achieve this, we first develop deadline- and energy-aware UBAC (DE-UBAC), an interpretable hierarchical rule-based realization that incorporates deadline-aware prioritization, cooperative placement, energy-aware coordination, and cache-update cost. However, predefined priorities and thresholds limit its adaptability under dynamically coupled network conditions. We therefore propose GRU-assisted Deadline- and Energy-aware Proximal Policy Optimization (GRU-DEPPO), which combines local and global GRU-based temporal representations with a centralized PPO actor--critic for adaptive cache placement and replacement. Simulation results show that GRU-DEPPO achieves the lowest deadline-miss rate and total UAV-side communication energy among the considered schemes. Compared with DE-UBAC, GRU-DEPPO reduces the deadline-miss rate by approximately 61\% and inter-UAV energy imbalance by approximately 40\%. 
\end{abstract}

\begin{keywords}
Unmanned aerial vehicles \sep Edge caching \sep Dynamic content demand \sep Deadline-aware caching \sep Energy-aware caching \sep Deep reinforcement learning
\end{keywords}

\maketitle
\pagestyle{plain}
\thispagestyle{plain}

\section{Introduction}
\label{sec:introduction}
Recent advances in the computational and sensing capabilities of end
devices, coupled with the rapid proliferation of mobile and Internet of
Things (IoT) devices, have led to a substantial increase in the volume of
data generated at the network edge~\citep{EdgeNetwork,EdgeRequireSurvey}.
At the same time, emerging applications have imposed increasingly diverse
and stringent requirements in terms of latency, reliability, and bandwidth.
Conventional cloud computing, which provides large-scale computing resources
through centralized data centers, faces several limitations in satisfying
these requirements: (i) communication latency caused by geographical distance
and the traversal of multiple network segments, (ii) latency variations and
jitter resulting from network congestion, (iii) increasing backhaul traffic,
and (iv) degraded service availability under disruptions in wide-area network
connectivity. To address these limitations, edge computing has attracted
considerable attention as a distributed computing paradigm that places
computing and storage resources closer to end devices and data sources~\citep{EdgeResourceDemand,rfc9556}. By processing data at or near the
network edge, edge computing can reduce service latency and alleviate the
traffic burden on backhaul networks.

In particular, unmanned aerial vehicle (UAV)-assisted edge computing has
attracted significant attention because it enables communication, storage,
and computing resources to be rapidly and flexibly deployed in environments
where fixed base stations and edge infrastructure are unavailable,
insufficient, or unable to respond promptly to abrupt changes in service
demand~\citep{UAV-enabled,UAVCDN,UAVCDN2}. Owing to their controllable
mobility and adjustable altitude, UAVs can be rapidly deployed and
repositioned to increase the probability of establishing line-of-sight (LoS)
links with ground devices and dynamically adjust communication coverage.
They can also accommodate temporary surges in traffic demand and provide
temporary network services in areas where terrestrial communication
infrastructure is sparse, unavailable, or damaged. These characteristics
make UAV-assisted edge computing particularly useful in remote environments,
such as mountainous and maritime regions; temporarily crowded venues, such
as concerts and sporting events; and disaster-affected areas where terrestrial
base stations or wired backhaul links have been damaged by earthquakes or
wildfires~\citep{UAVCDN3}.

Despite these advantages, UAV-assisted networks face several inherent
limitations, including energy constraints imposed by limited battery
capacity, restricted onboard computing and storage resources, and limited
backhaul capacity between UAVs and terrestrial networks~\citep{UAVDrawback}. In particular, forwarding user requests to a remote
content server or central cloud through a macro base station (MBS) may
increase backhaul traffic and content delivery latency. The associated
backhaul transmissions may also consume substantial communication energy,
thereby accelerating battery depletion and shortening the operational
lifetime of UAVs.
A promising approach to reducing backhaul dependence, content delivery
latency, and communication energy consumption is to equip UAVs with storage
and employ them as aerial cache nodes that proactively store selected
content~\citep{cacheenableUAV}. When the requested content is available in the local cache of the serving UAV, the UAV can deliver it directly to the user without accessing the MBS or retrieving the content from a remote server. If the content is
not cached locally but is available at a neighboring UAV, it can instead be
obtained through cooperative content sharing, further reducing reliance on the terrestrial backhaul. Such local and cooperative cache hits can reduce
content delivery latency, backhaul utilization, communication energy
consumption, and overall network resource usage.

However, because UAVs have limited storage capacity, the entire content
catalog cannot be cached. Moreover, content-demand and user-mobility patterns may vary across
different regions and over time, while different content types may have
heterogeneous delivery deadlines. A cache miss may therefore require content retrieval through the terrestrial backhaul, incurring additional delay and communication
energy consumption and potentially causing the request to exceed its deadline~\citep{DeadlineAware,DeadlineAware2}. Cache replacement also entails a non-negligible cost because fetching and storing new content requires
additional data transmission and communication energy. Furthermore, uneven
service loads across UAVs may result in imbalanced communication-energy
consumption.
Consequently, maximizing the cache hit ratio or selecting content solely
according to long-term popularity may be insufficient in dynamic multi-UAV
environments~\citep{Coop_caching,COCFG,ContentDemandPattern}. An effective caching strategy should therefore adapt cache placement to changes in content demand and network conditions while jointly reducing the deadline-miss rate, the communication energy consumed by request delivery and cache updates, and inter-UAV energy imbalance~\citep{EnergyBalance}. These improvements can also reduce dependence on MBS-assisted content retrieval.

To address these challenges, we first develop an interpretable deadline- and energy-aware rule-based caching design and then propose GRU-assisted Deadline- and Energy-aware PPO (GRU-DEPPO) to overcome the limitations of predefined cache-decision rules under dynamic multi-UAV conditions. GRU-DEPPO periodically determines cache placement and replacement decisions by jointly considering content-specific delivery deadlines, the communication energy incurred by request delivery and cache updates, and the distribution of communication energy consumption across UAVs. By using local and global GRUs to capture temporal patterns from UAV-local and network-wide observations, the proposed scheme adapts to spatially heterogeneous and time-varying network conditions while reducing deadline misses, communication energy consumption, and inter-UAV energy imbalance.

The main contributions of this study are as follows:
\begin{itemize}

\item \textbf{Dynamic deadline- and energy-aware multi-UAV caching formulation:}
We formulate a dynamic cache placement and replacement problem for multi-UAV edge networks with spatially heterogeneous and time-varying content demand and content-specific delivery deadlines. The formulation jointly considers the deadline-miss rate, UAV-side communication energy incurred by request delivery and cache updates, and inter-UAV communication-energy imbalance, subject to cache-capacity and cache-update feasibility constraints.

\item \textbf{Deadline- and energy-aware cache decision design:}
We develop DE-UBAC as an interpretable hierarchical realization of the
formulated caching problem. It extends user-behavior-aware caching with
deadline-miss-aware prioritization, persistent-miss feedback,
deadline-feasible cooperative placement, energy-aware cache coordination,
and cache-update-cost-aware replacement. By assigning these factors distinct roles in the
cache-decision process, DE-UBAC provides a structured rule-based
realization of the proposed problem and reveals the limitations of
predefined decision rules under dynamic multi-UAV conditions.

\item \textbf{GRU-assisted adaptive caching policy:}
To overcome the limitations of the above rule-based design, we develop GRU-DEPPO as the learning-based solution to adaptively optimize cache placement and replacement under dynamically coupled network conditions. A shared local
GRU captures UAV-specific temporal evolution, while a global GRU encodes
network-wide temporal context at the MBS. Their representations are fused
within a centralized PPO actor--critic that jointly learns feasible cache
updates from deadline, demand, connectivity, and energy conditions,
avoiding the fixed priorities and thresholds required by the rule-based
design.

\end{itemize}
The remainder of this paper is organized as follows. Section~\ref{sec:related_work} reviews related work on UAV-assisted edge
networks and edge caching. Section~\ref{sec:system_model} presents the system model and formulates the deadline- and energy-aware cache placement and replacement problem. Section~\ref{sec:gru_deppo} first develops a deadline- and energy-aware rule-based extension of UBAC and discusses its limitations under dynamic network conditions, and then presents the proposed GRU-DEPPO scheme for learning adaptive cache placement and replacement decisions. Section~\ref{sec:results} evaluates the proposed approaches under various experimental settings and compares them with baseline schemes. Finally, Section~\ref{sec:conclusion} concludes the paper and discusses directions for future research.

\section{Related Work}
\label{sec:related_work}

\begin{table*}[t]
\centering
\caption{Comparison of representative caching approaches for multi-UAV
edge networks.}
\label{tab1}
\renewcommand{\arraystretch}{1.15}
\setlength{\tabcolsep}{3pt}
\footnotesize

\begin{tabular*}{\textwidth}{
@{\extracolsep{\fill}}
l
l
c
c
c
c
c
c
c
}
\toprule
\makecell[rt]{Study}
&
\makecell[rt]{Type}
&
\makecell[c]{Multi-\\UAV}
&
\makecell[c]{Local\\awareness}
&
\makecell[c]{Inter-UAV\\cooperation}
&
\makecell[c]{Communication\\energy\\consumption}
&
\makecell[c]{Inter-UAV\\energy\\imbalance}
&
\makecell[c]{Content-specific\\deadline}
&
\makecell[c]{Realistic\\user pattern}
\\
\midrule

\citep{functionbased1,functionbased2,UBAC}
&
Utility- and rule-based
&
$\triangle$
&
$\triangle$
&
$\triangle$
&
$\triangle$
&
$\times$
&
$\times$
&
$\triangle$
\\

\citep{COCFG,Game2,Game3}
&
Game-theoretic
&
$\triangle$
&
$\times$
&
$\triangle$
&
$\triangle$
&
$\times$
&
$\times$
&
$\triangle$
\\

\citep{DRL1,DRL2,DRL3}
&
DRL-based
&
$\triangle$
&
$\times$
&
$\times$
&
$\triangle$
&
$\times$
&
$\times$
&
$\triangle$
\\

\citep{DRL4}
&
DL-assisted DRL
&
$\bigcirc$
&
$\triangle$
&
$\times$
&
$\times$
&
$\times$
&
$\times$
&
$\bigcirc$
\\

GRU-DEPPO
&
DL-assisted DRL
&
$\bigcirc$
&
$\bigcirc$
&
$\bigcirc$
&
$\bigcirc$
&
$\bigcirc$
&
$\bigcirc$
&
$\bigcirc$
\\

\bottomrule
\end{tabular*}

\vspace{1mm}
\parbox{\textwidth}{
\centering
\footnotesize
$\bigcirc$: supported;
$\triangle$: partially supported;
$\times$: not supported.

For grouped studies, $\triangle$ indicates that the corresponding feature
is supported only by a subset of the included studies or is considered in
a limited manner.
}
\end{table*}

This section reviews existing approaches for serving diverse user requests
in multi-UAV networks; their key characteristics are summarized in
Table~\ref{tab1}.

Traditional cache replacement policies, such as Least Frequently Used (LFU)
and Least Recently Used (LRU), make cache decisions using simple frequency-
or recency-based statistics and therefore incur low computational overhead.
However, their fixed decision rules do not explicitly account for dynamic
network conditions, inter-UAV cooperation, content-specific deadlines, or
the communication and cache-update energy costs. Several studies have incorporated additional network information into cache placement decisions. The work in~\citep{functionbased1} jointly optimized caching probabilities and V2V transmission power to improve energy efficiency in a UAV-assisted IoV network, whereas~\citep{functionbased2} jointly optimized UAV deployment, cache placement, and user association to improve user QoE. The scheme in~\citep{UBAC} further incorporated localized request patterns and predicted user mobility into proactive cache updates. Although these approaches exploit richer network and user information than conventional replacement policies, they rely on predefined optimization or placement rules and do not jointly consider cache-update energy, inter-UAV energy imbalance, and content-specific deadline misses.

Game-theoretic approaches have also been used to coordinate caching and
resource allocation in UAV-assisted networks. The study in~\citep{COCFG}
modeled cooperative content transmission among UAVs using a coalition
formation game, while~\citep{Game2} employed a Stackelberg--Bertrand game
to jointly consider content caching and service economics. The work
in~\citep{Game3} considered content caching, service placement, and task
offloading using Gibbs sampling and a matching game. These approaches
provide structured mechanisms for coordinating multiple network entities,
but they generally rely on predefined utility or preference models and do
not learn temporal demand dynamics from historical observations. Moreover,
they do not jointly address cache-update energy, inter-UAV energy imbalance,
and request-level deadline misses under content-specific deadlines.

Learning-based approaches have been investigated to improve adaptability in
dynamic UAV networks. The work in~\citep{DRL1} combined differential
evolution for content caching with multi-agent PPO for UAV trajectory
optimization. In~\citep{DRL2}, a dueling DQN jointly determined UAV
positions and cache placement using user-location and request information,
whereas~\citep{DRL3} employed dual-clip PPO to jointly optimize caching,
trajectory, and transmission power in a single-UAV setting. More recently,
the work in~\citep{DRL4} used an Informer model to predict user request and
mobility patterns and incorporated these predictions into a PPO-based
deployment framework. These learning-based approaches improve adaptability to dynamic user and network conditions, but their objectives primarily focus on average delay, QoE, trajectory, or transmission power. They do not jointly model the
communication energy consumed by request delivery and cache updates, inter-UAV energy imbalance, and content-specific request deadline misses. Furthermore, existing learning-based schemes do not directly encode recent UAV-local network states to capture temporal dynamics for adaptive cache-update decisions.

In summary, existing caching schemes typically address only a subset of
local demand awareness, dynamic demand adaptation, cooperative caching,
communication energy, and heterogeneous delivery requirements. In contrast, this study makes periodic
cache-update decisions by jointly considering content-specific deadline misses
at the request level, communication energy from both request delivery and
cache updates, and inter-UAV energy imbalance under time-varying demand.

\section{System Model and Problem Formulation}
\label{sec:system_model}

\begin{figure}[pos=t]
\centering
\includegraphics[width=0.48\textwidth]{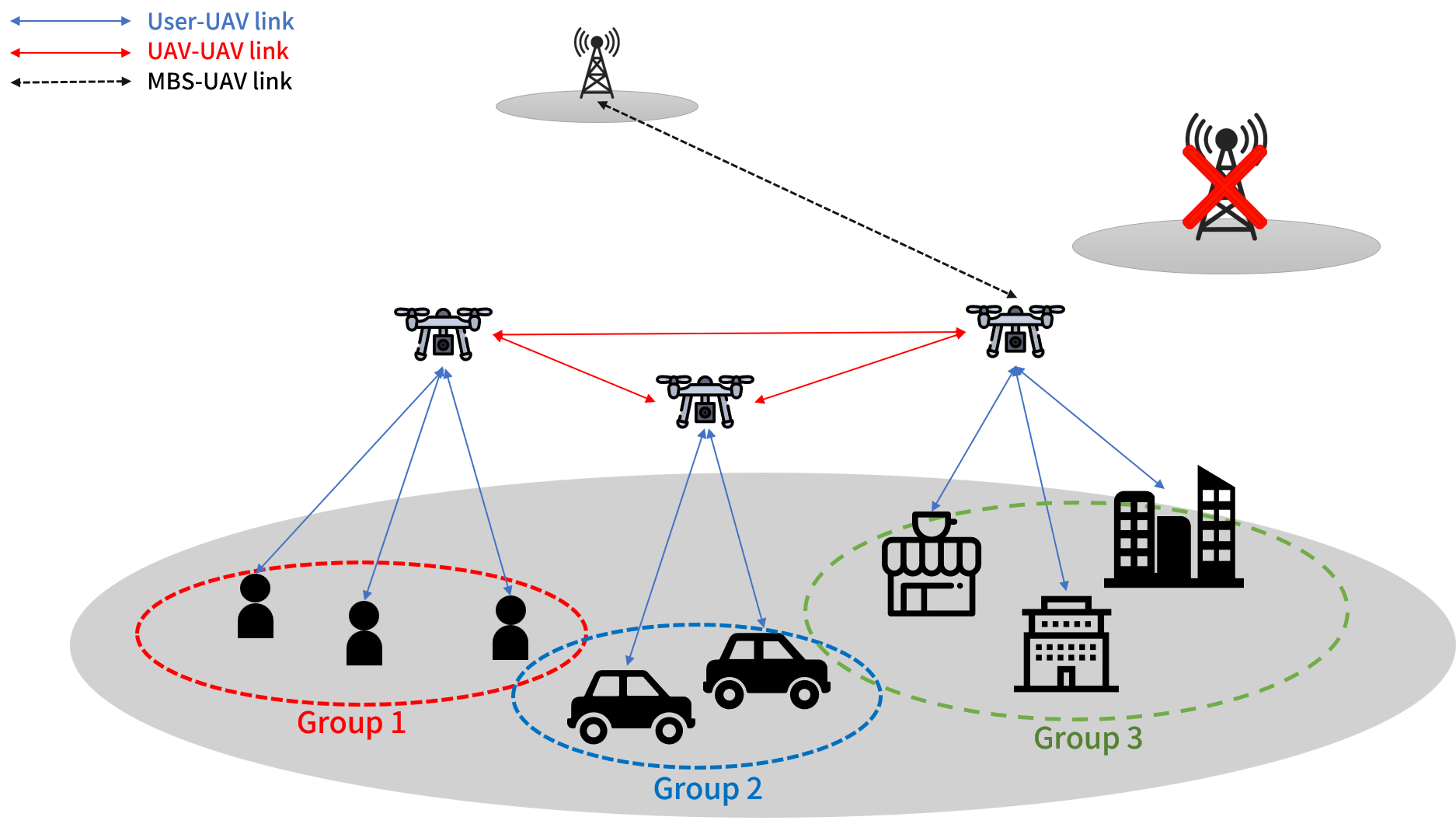}
\caption{System model of the considered UAV network}
\label{fig:system_model}
\end{figure}

In the considered scenario illustrated in Fig.~\ref{fig:system_model}, the local base station is unavailable due to a failure, and a remote MBS stores the global content pool and provides backhaul access to cache-enabled UAVs. Users are divided into groups, with each group characterized by group-specific mobility patterns and content-request distributions that are shared by its users. Each user associates with the nearest UAV having a feasible access link. Requests may be served by the associated UAV, a neighboring UAV, or the MBS, depending on content availability and link feasibility~\citep{cooperative}. The selected path determines the request delay and communication energy consumption.

Users move within the service area and generate requests according to group-level preferences and request-generation probabilities. Each generated request follows a strict source-selection priority. For a request associated with UAV $n$, the local cache is checked first. If the requested item is not available locally, UAV $n$ searches for feasible cooperative paths to UAVs caching the item and selects the path with the smallest end-to-end transmission delay. Only when no feasible cooperative cache path exists is the item retrieved from the MBS through the feasible direct or relayed backhaul path with the smallest end-to-end transmission delay. If no feasible delivery path is available owing to link outages or insufficient achievable rates, the corresponding request is dropped and excluded from the delivered-request set.

Cache placement is updated at predefined intervals based on the observed network state. Newly cached content is transferred either directly from the MBS to the target UAV or through one or more UAV relay hops. To model nonstationarity, a change point is randomly selected from the latter half of each episode, at which both the mobility context and local request distributions shift.

Let $\mathcal{N}$, $\mathcal{U}$, $\mathcal{C}$, $\mathcal{T}$, and
$\mathcal{R}^{\mathrm{all}}$ denote the UAV, user, content, content-type,
and episode-level request sets, respectively. The discrete simulation-time
index and cache-decision interval duration are denoted by $t$ and
$\Delta_c$, respectively. For each request $k\in\mathcal{R}^{\mathrm{all}}$,
let $\Pi_k$ denote its selected delivery path, with
$\Pi_k=\varnothing$ when no feasible path is
available. The set of successfully delivered requests is therefore
$\mathcal{R}^{\mathrm{srv}}=\{k\in\mathcal{R}^{\mathrm{all}}:
\Pi_k\neq\varnothing\}$.
 
\subsection{Transmission Model}
We use large-scale propagation models for the User--UAV, inter-UAV, and MBS--UAV links~\citep{3gpp,6863654,channelmodel1,channelmodel2}. The User--UAV link additionally accounts for elevation-dependent LoS/NLoS attenuation. The achievable rate of each link is determined by its available bandwidth and SINR. A link is considered infeasible if its SINR falls below $\Gamma_{\min}$ or its achievable rate falls below $R_{\min}$. Cache updates use either a direct MBS--UAV link or an MBS--UAV link followed by one or more UAV--UAV relay links.

\subsubsection{User--UAV Access Link}

The link between UAV $n$ and user $u$ is modeled as an air-to-ground
wireless link with three-dimensional distance $d_{n,u}$ and elevation angle
$\theta_{n,u}$.

The probability of an LoS connection is modeled as an
elevation-angle-dependent function~\citep{6863654}:
\begin{equation}
\label{eq:los}
\makebox[\linewidth][c]{%
$\displaystyle
\begin{gathered}
P_{n,u}^{\mathrm{LoS}}
=
\frac{1}
{1+X\exp[-Y(\theta_{n,u}-X)]},
\\[3pt]
P_{n,u}^{\mathrm{NLoS}}
=
1-P_{n,u}^{\mathrm{LoS}}
\end{gathered}
$%
}
\end{equation}
Here, $X$ and $Y$ are environment-dependent parameters. The average path
loss is
\begin{equation}
\label{eq:avgpathloss}
\makebox[\linewidth][c]{%
$\displaystyle
\begin{gathered}
PL_{n,u}^{\mathrm{FS}}
=
20\log_{10}
\left(
\frac{4\pi f_u d_{n,u}}{c}
\right),
\\[3pt]
PL_{n,u}
=
PL_{n,u}^{\mathrm{FS}}
+
P_{n,u}^{\mathrm{LoS}}\eta_{\mathrm{LoS}}
+
P_{n,u}^{\mathrm{NLoS}}\eta_{\mathrm{NLoS}}
\end{gathered}
$%
}
\end{equation}
where $f_u$ is the carrier frequency, $c$ is the speed of light, and
$\eta_{\mathrm{LoS}}$ and $\eta_{\mathrm{NLoS}}$ are the additional LoS
and NLoS losses. The SINR, rate, and transmission delay are
\begin{equation}
\label{eq:access_performance}
\makebox[\linewidth][c]{%
$\displaystyle
\begin{gathered}
\Gamma_{n,u}
=
\frac{
P_n^{\mathrm{tx}}10^{-PL_{n,u}/10}
}{
N_0B_{n,u}^{\mathrm{eff}}+I_{n,u}
},
\\[3pt]
R_{n,u}
=
B_{n,u}^{\mathrm{eff}}
\log_2(1+\Gamma_{n,u}),
\\[3pt]
D_{n,u}^{\mathrm{tx}}
=
\frac{S}{R_{n,u}}
\end{gathered}
$%
}
\end{equation}
Here, $P_n^{\mathrm{tx}}$, $N_0$, $B_{n,u}^{\mathrm{eff}}$, and $I_{n,u}$
denote the transmit power, noise power spectral density, effective
bandwidth, and aggregate co-channel interference, respectively. The payload
size is denoted by $S$.
\subsubsection{UAV--UAV Relay Link}
The relay link between UAVs $n$ and $m$ is modeled using a
distance-dependent free-space path-loss model:
\begin{equation}
\label{eq:uav_uav_pathloss}
\makebox[\linewidth][c]{%
$\displaystyle
PL_{n,m}
=
20\log_{10}
\left(
\frac{4\pi f_{\mathrm{UAV}}d_{n,m}}{c}
\right)
$%
}
\end{equation}
where $d_{n,m}$ denotes the distance between UAVs $n$ and $m$, and
$f_{\mathrm{UAV}}$ is the carrier frequency of the inter-UAV link.
The corresponding SINR, achievable transmission rate, and transmission
delay are calculated using~\eqref{eq:access_performance}, with the
inter-UAV transmit power, bandwidth, and interference terms. When multi-hop
relaying is employed, the delay and energy consumption of the individual
hops are accumulated over the complete relay path.

\subsubsection{MBS--UAV Backhaul Link}
The backhaul link between the MBS, indexed by $0$, and UAV $n$ is modeled
using a distance-dependent free-space path-loss model with an additional
attenuation term:
\begin{equation}
\label{eq:mbs_uav_pathloss}
\makebox[\linewidth][c]{%
$\displaystyle
PL_{0,n}
=
20\log_{10}
\left(
\frac{4\pi f_{\mathrm{MBS}}d_{0,n}}{c}
\right)
+
\eta_{\mathrm{MBS}}
$%
}
\end{equation}
where $d_{0,n}$ denotes the distance between the MBS and UAV $n$,
$f_{\mathrm{MBS}}$ is the carrier frequency of the MBS--UAV link, and
$\eta_{\mathrm{MBS}}$ denotes the additional propagation attenuation.
The corresponding SINR, achievable transmission rate, and transmission
delay are calculated using~\eqref{eq:access_performance}, with the MBS
transmit power, backhaul bandwidth, and interference terms.

\subsection{Content Request Model}
\label{sub:crm}
This subsection defines type-specific deadlines and time-varying, user-dependent content requests.

\subsubsection{Content Characteristics}
Each content item $i$ has size $S_i$, type $\tau_i$, and delivery deadline
$\delta_i$. For modeling tractability and to isolate the effects of caching
decisions, all items are assumed to have an identical size $S$, i.e.,
$S_i=S$ for all $i\in\mathcal{C}$. The content catalog is partitioned into
type-specific subsets, each associated with a different deadline:
\begin{equation}
\label{eq:type_deadline}
\makebox[\linewidth][c]{%
$\displaystyle
\delta_i
=
\delta_{\tau_i}
$%
}
\end{equation}
Thus, requests for different content types may have different
delivery-time requirements.

\subsubsection{Dynamic Request Generation}

At each time step, requests follow a Zipf-like global popularity
distribution~\citep{zipf,zipf2,zipf3}:
\begin{equation}
\label{eq:global_popularity}
\makebox[\linewidth][c]{%
$\displaystyle
p_i^{\mathrm{global}}
=
\frac{q_i^{-\alpha}}
{\sum_{j\in\mathcal{C}}q_j^{-\alpha}}
$%
}
\end{equation}
where $\mathcal{C}$ denotes the content catalog, $q_i$ is the
popularity rank of content $i$, and $\alpha$ is the Zipf parameter.

Motivated by group-level heterogeneous demand modeling
in~\citep{UBAC}, we perturb the global distribution according to
group-specific preferences, spatial regions, and time-varying demand states.
For group $g$ at time $t$, the resulting local popularity is defined as
\begin{equation}
\label{eq:local_popularity}
\makebox[\linewidth][c]{%
$\displaystyle
p_{i,g,t}
=
\frac{
p_i^{\mathrm{global}}
\exp\left(\epsilon_{i,g,t}\right)
}{
\sum_{j\in\mathcal{C}}
p_j^{\mathrm{global}}
\exp\left(\epsilon_{j,g,t}\right)
}
$%
}
\end{equation}
where $\epsilon_{i,g,t}$ is a group- and time-dependent content-specific perturbation.

Let $A_{u,t}$ indicate whether user $u$ generates a request at time $t$:
\begin{equation}
\label{eq:request_generation}
\makebox[\linewidth][c]{%
$\displaystyle
\begin{gathered}
A_{u,t}\sim\mathrm{Bernoulli}(\lambda_{u,t}),
\\[3pt]
J_{u,t}\sim
\mathrm{Categorical}
\left(
\left\{
p_{i,g(u),t}
\right\}_{i\in\mathcal{C}}
\right)
\quad\text{if } A_{u,t}=1
\end{gathered}
$%
}
\end{equation}
where $\lambda_{u,t}$ is the request-generation probability and
$J_{u,t}$ is the requested content.

To capture nonstationarity, a change point is selected from the second half
of each episode, at which the local distribution shifts toward a newly
sampled demand state:
\begin{equation}
\label{eq:demand_shift}
\makebox[\linewidth][c]{%
$\displaystyle
p_{i,g,t}^{\mathrm{new}}
=
(1-\beta_{\mathrm{shift}})p_{i,g,t}^{\mathrm{old}}
+
\beta_{\mathrm{shift}}\widetilde{p}_{i,g,t}
$%
}
\end{equation}
where $\widetilde{p}_{i,g,t}$ is the newly sampled distribution and
$\beta_{\mathrm{shift}}\in[0,1]$ controls the shift magnitude.

\subsection{Energy Consumption Model}

We consider the UAV-side communication energy associated with content
delivery and cache updates~\citep{ener1}. For a transfer of content item
$i$ over link $\ell$, let $v_{\ell}^{\mathrm{tx}}$ and
$v_{\ell}^{\mathrm{rx}}$ denote the transmitting and receiving nodes,
respectively. The UAV-side communication energy incurred over link
$\ell$ is defined as
\begin{equation}
\label{eq:link_energy}
\makebox[\linewidth][c]{%
$\displaystyle
E_{\ell}
=
\left(
\mathbb{1}\{v_{\ell}^{\mathrm{tx}}\in\mathcal{N}\}
P_{\ell}^{\mathrm{tx}}
+
\mathbb{1}\{v_{\ell}^{\mathrm{rx}}\in\mathcal{N}\}
P_{\ell}^{\mathrm{rx}}
\right)
\frac{S_i}{R_{\ell}}
$%
}
\end{equation}
where $P_{\ell}^{\mathrm{tx}}$ and $P_{\ell}^{\mathrm{rx}}$ denote the
transmit-side and receive-side power draws, respectively, and
$R_{\ell}$ is the link rate. Thus, only the energy consumed by UAVs is
included, while the communication energy consumed by the MBS and users
is excluded.

Let $\mathcal{R}^{\mathrm{srv}}$ denote the set of successfully
delivered requests and let $\mathcal{Q}$ denote the link-level
transmissions used for cache updates. The total UAV-side communication
energy is
\begin{equation}
\label{eq:total_energy}
\makebox[\linewidth][c]{%
$\displaystyle
E^{\mathrm{total}}
=
\sum_{k\in\mathcal{R}^{\mathrm{srv}}}
\sum_{\ell\in\Pi_k}E_{\ell}
+
\sum_{q\in\mathcal{Q}}E_q
$%
}
\end{equation}
The first term captures the UAV-side communication energy incurred by
request delivery, including multi-hop relay transmissions, and the
second captures the UAV-side communication energy incurred by cache
updates. Cache evictions are local and incur negligible communication
energy.

Let $E_n$ denote the communication energy attributed to UAV $n$ during
the episode. For each link-level transmission, the transmit-side energy
is attributed to UAV $n$ when it acts as the transmitter, whereas the
receive-side energy is attributed to UAV $n$ when it acts as the
receiver. Accordingly,
\begin{equation}
E^{\mathrm{total}}
=
\sum_{n\in\mathcal{N}} E_n .
\end{equation}

The resulting inter-UAV communication-energy imbalance is evaluated using the coefficient
of variation:
\begin{equation}
\label{eq:energy_imbalance}
\makebox[\linewidth][c]{%
$\displaystyle
\mathcal{I}_{E}
=
\frac{
\sqrt{
\frac{1}{N}
\sum_{n=1}^{N}
\left(E_n-\overline{E}\right)^2
}
}{
\overline{E}+\epsilon_E
},
\qquad
\overline{E}
=
\frac{1}{N}
\sum_{n=1}^{N}E_n
$%
}
\end{equation}
where $N$ is the number of UAVs and $\epsilon_E$ is a small positive
constant. A lower value of $\mathcal{I}_{E}$ indicates more balanced
communication energy consumption across UAVs.

\subsection{Deadline Metric, Cache Constraints, and Problem Formulation}
This subsection defines the deadline metric and cache constraints and formulates a
joint optimization problem. Using the models introduced above, the objective
is to jointly reduce the deadline-miss rate, total communication energy,
and inter-UAV energy imbalance.

\subsubsection{Deadline-Miss Metric}

For request $k\in\mathcal{R}^{\mathrm{all}}$, let $i_k$ denote the
requested content. Its deadline is $\delta_k=\delta_{\tau_{i_k}}$. For a
request with a feasible path, its end-to-end delay is
\begin{equation}
\label{eq:request_delay}
\makebox[\linewidth][c]{%
$\displaystyle
D_k=\sum_{\ell\in\Pi_k}D_{\ell}^{\mathrm{tx}}
$%
}
\end{equation}
Here, $D_{\ell}^{\mathrm{tx}}$ is obtained from the corresponding link
model. Resource contention is reflected through the effective bandwidth and
interference terms in the link model; an additional queuing delay is not
explicitly modeled.

For each delivered request, a deadline miss occurs when its end-to-end delay
exceeds its content-specific deadline:
\begin{equation}
\label{eq:deadline_metrics}
\makebox[\linewidth][c]{%
$\displaystyle
\begin{gathered}
M_k=\mathbb{1}\{D_k>\delta_k\},
\qquad k\in\mathcal{R}^{\mathrm{srv}},
\\[3pt]
\mathrm{DMR}=
\frac{\sum_{k\in\mathcal{R}^{\mathrm{srv}}}M_k}
{\max\left(1,|\mathcal{R}^{\mathrm{srv}}|\right)}
\end{gathered}
$%
}
\end{equation}
Here, $\mathrm{DMR}$ is the deadline-miss rate among delivered requests.

\subsubsection{Cache Constraints}
\label{subsub:cacheconstraints}

Let $x_{n,i}(t)\in\{0,1\}$ indicate whether UAV $n$ caches item $i$ at
time $t$. The cache state is constrained by
\begin{equation}
\label{eq:cache_constraint}
\makebox[\linewidth][c]{%
$\displaystyle
\begin{gathered}
x_{n,i}(t),\,\widehat{x}_{n,i}(t)\in\{0,1\},
\\[3pt]
\sum_{i\in\mathcal{C}}x_{n,i}(t)\leq K_n,
\qquad
\sum_{i\in\mathcal{C}}\widehat{x}_{n,i}(t)\leq K_n,
\qquad \forall n\in\mathcal{N}
\end{gathered}
$%
}
\end{equation}
where $K_n$ is the cache capacity in items. At each cache-update instant,
the controller selects a target cache state
$\widehat{x}_{n,i}(t)$ that satisfies this capacity constraint.

Let $\mathcal{C}_{n}^{+}(t)=\{i:\widehat{x}_{n,i}(t)=1,\,x_{n,i}(t)=0\}$ denote
the newly inserted items. The target state is not activated immediately:
the previous cache state remains available while the required backhaul
transfers are performed. Let $\varphi_n(t)$ denote the selected cache-update
path from the MBS, indexed by $0$, to UAV $n$. A direct update contains only the MBS--UAV link, whereas a relayed update contains that link followed by
one or more UAV--UAV links. Assuming sequential content additions, their
duration must satisfy
\begin{equation}
\label{eq:cache_update_feasibility}
\makebox[\linewidth][c]{
$\displaystyle
\begin{gathered}
D_{n,t}^{\mathrm{cache}}
=
\sum_{i\in\mathcal{C}_{n}^{+}(t)}
\sum_{\ell\in\varphi_n(t)}
\frac{S_i}{R_{\ell}(t)}
\leq
\Delta_c,
\\[3pt]
\forall n\in\mathcal{N}
\end{gathered}
$%
}
\end{equation}
where $R_{\ell}(t)$ is the rate of link $\ell$ at the update instant. After
this delay, the target state becomes active and the corresponding link-level
transmissions are included in $\mathcal{Q}$. A target cache state that
violates~\eqref{eq:cache_update_feasibility} is treated as an invalid action
and excluded from the controller's feasible action set. Cache evictions are
local operations.

\subsubsection{Problem Formulation}
Let $\mu$ denote a dynamic caching policy that maps the observed network
state to target cache states. We formulate the dynamic caching problem as
the following multi-objective optimization:
\begin{equation}
\label{eq:problem_formulation}
\makebox[\linewidth][c]{%
$\displaystyle
\min_{\mu}
\mathbb{E}\left[
\left(
\mathrm{DMR},
E^{\mathrm{total}},
\mathcal{I}_E
\right)
\right]
$%
}
\end{equation}
subject to~\eqref{eq:cache_constraint} and
\eqref{eq:cache_update_feasibility}. The objective jointly minimizes the
deadline-miss rate, total communication energy, and inter-UAV energy
imbalance.

Problem~\eqref{eq:problem_formulation} is a stochastic combinatorial optimization problem. Even when user locations, requests, and link conditions are fixed, the formulation contains a static cooperative cache-placement subproblem with binary placement decisions and per-UAV cache-capacity constraints. Cooperative cache-placement problems of this form have been shown to be NP-hard through a reduction from the 2-Disjoint Set Covers problem~\citep{NP-Hard}. Therefore, Problem~\eqref{eq:problem_formulation}, which further incorporates time-varying demand, user mobility, content-specific deadlines, communication energy consumption, and inter-UAV energy imbalance, is also NP-hard.

The number of feasible cache placements grows combinatorially with the network size, making exact offline optimization computationally prohibitive for large-scale networks. Moreover, repeatedly solving the problem at each cache-decision interval is impractical under dynamically changing requests, mobility, and link conditions. We therefore first investigate a structured rule-based realization of the formulated problem and then develop a learning-based policy to overcome the limitations of predefined cache-decision rules.

\begin{figure*}[t]
\centering
\includegraphics[width=0.75\linewidth]{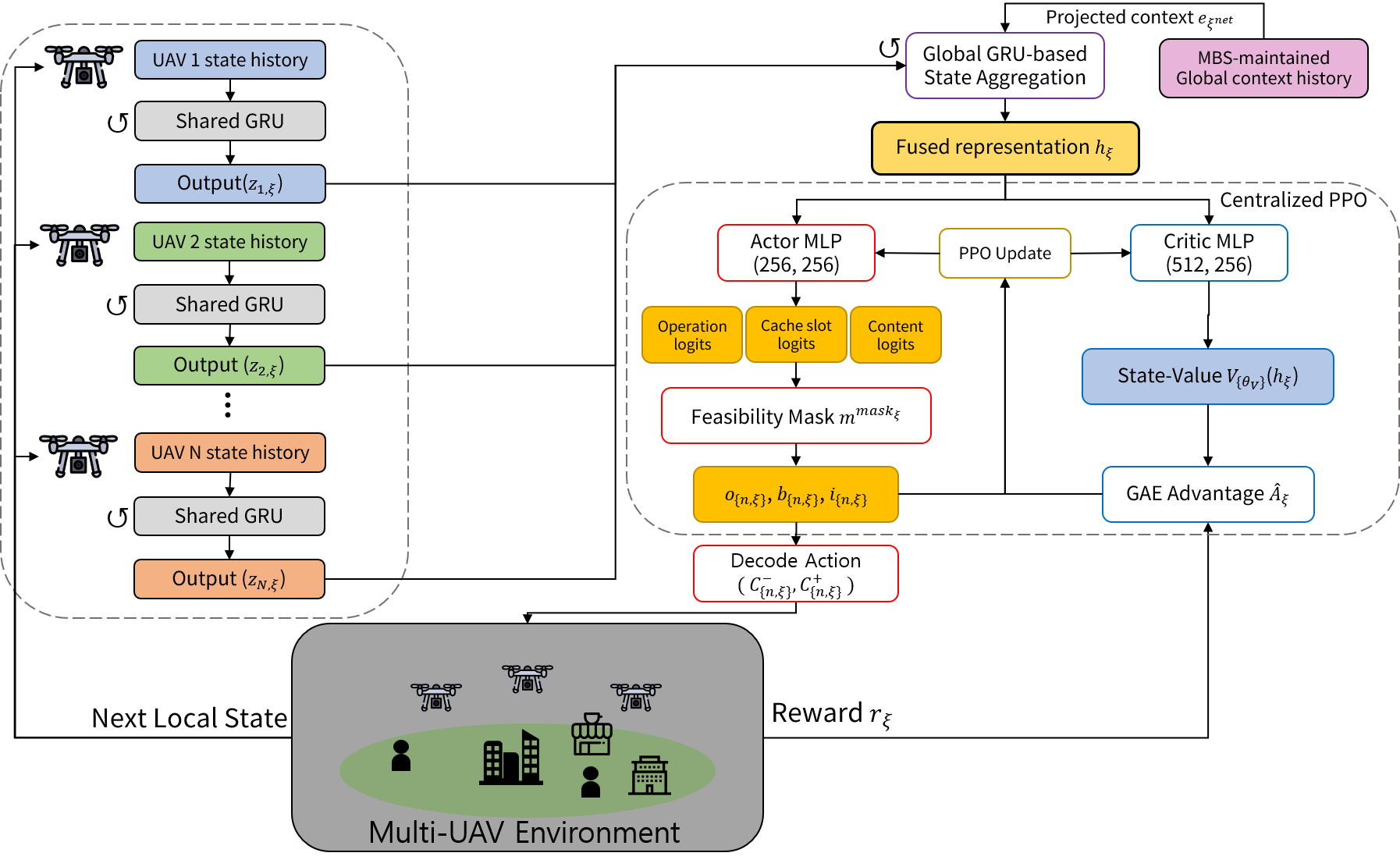}
\caption{The GRU-DEPPO scheme}
\label{fig:ppo_training_flow}
\end{figure*}

\section{Adaptive Deadline- and Energy-Aware Cache Algorithms}
\label{sec:gru_deppo}

In this section, we first investigate how the deadline- and energy-aware cache placement and replacement problem formulated in
Section~\ref{sec:system_model} can be addressed through a rule-based extension of the local-demand-aware UBAC scheme~\citep{UBAC}.
The resulting design incorporates content-specific deadlines, cooperative content accessibility, UAV communication-energy conditions, cache redundancy, and cache-update cost into cache placement and replacement decisions. We then discuss the limitations of relying on predefined rules under dynamically changing network conditions and formulate the cache placement and replacement process as a Markov decision process (MDP).
To capture the temporal evolution of spatially heterogeneous local and network-wide conditions, GRU-DEPPO employs local and global GRU-based representations, which are subsequently provided to a centralized PPO actor--critic for joint cache-update decision making. The main notation used throughout this section is summarized in Table~\ref{tab:notation_sec4}, and the overall GRU-DEPPO procedure is illustrated in Fig.~\ref{fig:ppo_training_flow}.

\begin{table*}[t]
\centering
\caption{Main notation used for DE-UBAC, the MDP, and GRU-DEPPO}
\label{tab:notation_sec4}
\renewcommand{\arraystretch}{1.12}
\setlength{\tabcolsep}{5pt}
\footnotesize

\begin{tabularx}{\textwidth}{
@{} l >{\raggedright\arraybackslash}X
l >{\raggedright\arraybackslash}X @{}
}
\toprule
Symbol & Description & Symbol & Description \\
\midrule

$\mathcal{N}$ &
Set of UAVs &
$\mathcal{C}$ &
Set of content items \\

$\xi$ &
Cache-decision epoch &
$\Delta_c$ &
Cache-decision interval \\

$C_k^{\mathrm{DE}}$ &
Deadline-aware counting increment for request $k$ &
$H_{n,i,\xi}^{\mathrm{DE}}$ &
Deadline-aware history count for UAV--content pair $(n,i)$ \\

$\widetilde{H}_{n,i,\xi}^{\mathrm{DE}}$ &
MBS-adjusted deadline-aware priority &
$p_{n,i,\xi}^{\mathrm{per}}$ &
Persistent deadline-miss indicator \\

$\mathcal{D}_{\xi}$ &
Set of nominated UAV--content pairs &
$\mathcal{F}_{n,i,\xi}^{\mathrm{dl}}$ &
Deadline-feasible cooperative UAV set \\

$\Delta P_{i,j}$ &
Normalized priority gain for replacing $j$ with $i$ &
$\widetilde{E}_{i}^{\mathrm{up}}$ &
Normalized communication energy for cache update \\

$s_{n,\xi}^{\mathrm{local}}$ &
Local state of UAV $n$ &
$s_{\xi}^{\mathrm{global}}$ &
Global network state \\

$\mathbf{x}_{n,\xi}$ &
Cache state of UAV $n$ &
$\mathbf{q}_{n,\xi}^{\mathrm{dem}}$ &
Local demand state of UAV $n$ \\

$\mathbf{g}_{n,\xi}^{\mathrm{conn}}$ &
Connectivity state of UAV $n$ &
$E_{n,\xi}^{\mathrm{comm}}$ &
Communication-energy state of UAV $n$ \\

$\boldsymbol{\psi}_{n,\xi}^{\mathrm{dl}}$ &
Deadline-pressure vector of UAV $n$ &
$a_\xi$ &
Joint cache-update action \\

$\mathcal{C}_{n,\xi}^{-}$, $\mathcal{C}_{n,\xi}^{+}$ &
Removed and added content sets &
$\mathbf{z}_{n,\xi}$ &
Local temporal representation \\

$\mathbf{z}_{\xi}^{\mathrm{global}}$ &
Global temporal representation &
$\mathbf{h}_{\xi}$ &
Fused temporal representation \\

$\mathbf{a}_{n,\xi}$ &
Cache-update action of UAV $n$ &
$\mathbf{m}_{\xi}^{\mathrm{mask}}$ &
Action feasibility mask \\

\bottomrule
\end{tabularx}
\end{table*}

\subsection{Deadline- and Energy-Aware Rule-Based Caching}
Caching decisions based primarily on content popularity or local demand can improve content availability around current request locations.
However, such decisions may be insufficient when contents have heterogeneous delivery deadlines and UAVs experience uneven communication-energy burdens. In particular, the preference for local cache service does not imply that every highly requested content should be replicated at its associated UAV. A content with a relatively relaxed deadline may still be served through a nearby cooperative cache within its deadline, allowing limited local cache space to be reserved for more deadline-sensitive contents. Moreover, repeatedly concentrating cache and service loads on particular UAVs may increase inter-UAV energy imbalance.

To incorporate these considerations into an interpretable rule-based strategy, we extend UBAC~\citep{UBAC} into a deadline- and energy-aware variant, referred to as DE-UBAC. DE-UBAC retains the user-behavior-aware counting and coordinated cache-management structure of UBAC, including handover-aware history updates and MBS-level cache coordination, while introducing deadline-, cooperation-, and energy-aware decision rules.

DE-UBAC employs a hierarchical decision process rather than combining all factors into a single weighted caching score. First, the UBAC counting process is extended to distinguish requests that satisfy their deadlines from those that miss them. For request $k$, the deadline-miss indicator is defined as
\begin{equation}
\makebox[\linewidth][c]{%
$\displaystyle
m_k^{\mathrm{dl}} = \mathbb{I}\!\left(D_k^{\mathrm{tar}}>\delta_{i_k}\right)
$%
}
\end{equation}
where $D_k^{\mathrm{tar}}$ denotes the delivery delay evaluated from the perspective of the target UAV and $\delta_{i_k}$ is the deadline of requested content $i_k$. The deadline-aware counting increment is given by
\begin{equation}
\makebox[\linewidth][c]{%
$\displaystyle
C_k^{\mathrm{DE}}
=
C_k^{\mathrm{UBAC}}
\left(1+\lambda_{\mathrm{dl}}m_k^{\mathrm{dl}}\right)
$%
}
\end{equation}
where $C_k^{\mathrm{UBAC}}$ denotes the original UBAC counting increment and $\lambda_{\mathrm{dl}}$ controls the additional weight assigned to a deadline-missed request. Consequently, deadline-missed requests accumulate caching priority more rapidly without differentiating the magnitude of the deadline violation.

DE-UBAC also accounts for deadline misses that persist after a previous cache adaptation. The MBS maintains a persistent-miss indicator $p_{n,i,\xi}^{\mathrm{per}}$ for each UAV--content pair and adjusts its deadline-aware priority as
\begin{equation}
\makebox[\linewidth][c]{%
$\displaystyle
\widetilde{H}_{n,i,\xi}^{\mathrm{DE}}
=
H_{n,i,\xi}^{\mathrm{DE}}
\left(1+\lambda_{\mathrm{per}}
p_{n,i,\xi}^{\mathrm{per}}\right)
$%
}
\end{equation}
when deadline misses remain unresolved after placement or relocation.
Here, $H_{n,i,\xi}^{\mathrm{DE}}$ denotes the accumulated deadline-aware history-table count, $p_{n,i,\xi}^{\mathrm{per}}\in\{0,1\}$ indicates a persistent deadline miss, and $\lambda_{\mathrm{per}}$ controls its additional priority.
This gives repeatedly unsatisfied UAV--content pairs stronger priority for subsequent cache adaptation.

Second, content placement and redundancy are determined according to deadline feasibility. If an existing cooperative copy can satisfy the content deadline, additional replication is avoided. Otherwise, DE-UBAC permits local or nearby placement to provide a shorter delivery path.
The target UAV is preferred whenever feasible; otherwise, a nearby UAV capable of satisfying the deadline is considered.

Third, DE-UBAC incorporates UAV communication-energy conditions as a secondary decision criterion. Deadline-aware priority remains dominant when candidate priorities are clearly different. When their normalized priority difference is below a predefined threshold $\tau_H$, the corresponding caching demands are regarded as comparable, and the UAV communication-energy burden is used to resolve the decision. If the target UAV cannot directly accommodate the content, a deadline-feasible alternative UAV with a relatively lower energy burden is preferred. This hierarchical rule allows energy balancing to influence caching decisions without overriding a clearly stronger deadline-aware demand.

Finally, DE-UBAC considers cache-update energy before applying a replacement. For candidate content $i$ and eviction candidate $j$, a replacement is performed only when
\begin{equation}
\makebox[\linewidth][c]{%
$\displaystyle
\Delta P_{i,j}
>
\lambda_{\mathrm{up}}
\widetilde{E}_{i}^{\mathrm{up}},
$%
}
\end{equation}
where $\Delta P_{i,j}$ denotes the normalized priority gain, $\widetilde{E}_{i}^{\mathrm{up}}$ denotes the normalized communication energy required for the cache update, and $\lambda_{\mathrm{up}}$ determines how strongly the cache-update energy penalizes replacement. Otherwise, the current cache configuration is retained.
Algorithm~\ref{alg:de_ubac} summarizes the overall DE-UBAC procedure.

\begin{algorithm}[t]
\caption{Deadline- and Energy-Aware UBAC}
\label{alg:de_ubac}
\begin{algorithmic}[1]

\Require $\{\mathbf{x}_{n,\xi}\}_{n\in\mathcal{N}}$,
$\{H_{n,i,\xi}^{\mathrm{DE}}\}$,
$\{\delta_i\}$, $\{E_{n,\xi}^{\mathrm{comm}}\}$
\Ensure Updated cache configuration

\For{each request $k$}
    \State $n \gets$ target UAV from the UBAC mobility rule
    \State $H_{n,i_k,\xi}^{\mathrm{DE}}
    \gets H_{n,i_k,\xi}^{\mathrm{DE}} + C_k^{\mathrm{DE}}$
\EndFor

\If{cache-decision epoch $\xi$}

    \Statex \textbf{Stage 1: Deadline-aware prioritization}
    \State Update $\{p_{n,i,\xi}^{\mathrm{per}}\}_{n,i}$
    \For{$n\in\mathcal{N},\,i\in\mathcal{C}$}
        \State $\widetilde H_{n,i,\xi}^{\mathrm{DE}}
        \gets H_{n,i,\xi}^{\mathrm{DE}}
        (1+\lambda_{\mathrm{per}}p_{n,i,\xi}^{\mathrm{per}})$
    \EndFor
    \State Construct $\mathcal{D}_{\xi}$

    \Statex \textbf{Stage 2: Deadline-feasible placement}
    \For{each $(n,i)\in\mathcal{D}_{\xi}$}
        \State Construct $\mathcal{F}_{n,i,\xi}^{\mathrm{dl}}$
        \If{$i$ is available through $\mathcal{F}_{n,i,\xi}^{\mathrm{dl}}$}
            \State $\mathcal{D}_{\xi}\gets
            \mathcal{D}_{\xi}\setminus\{(n,i)\}$
        \EndIf
    \EndFor

    \Statex \textbf{Stage 3: Energy-aware coordination}
    \For{each competing placement for $i$}
        \If{normalized priority difference $>\tau_H$}
            \State Select larger $\widetilde H_{n,i,\xi}^{\mathrm{DE}}$
        \Else
            \State Retain demand with larger $E_{n,\xi}^{\mathrm{comm}}$
        \EndIf
    \EndFor

    \Statex \textbf{Stage 4: Update-cost-aware replacement}
    \For{each selected $(n,i)$}
        \State $j\gets$ UBAC eviction candidate
        \If{$\Delta P_{i,j}>
        \lambda_{\mathrm{up}}\widetilde E_i^{\mathrm{up}}$}
            \State $\mathcal{C}_{n,\xi}^{-}\gets\{j\}$,
            $\mathcal{C}_{n,\xi}^{+}\gets\{i\}$
        \Else
            \State $\mathcal{C}_{n,\xi}^{-}
            =\mathcal{C}_{n,\xi}^{+}=\varnothing$
        \EndIf
    \EndFor

    \State Apply
    $\{(\mathcal{C}_{n,\xi}^{-},
    \mathcal{C}_{n,\xi}^{+})\}_{n\in\mathcal{N}}$

\EndIf

\end{algorithmic}
\end{algorithm}
Although DE-UBAC provides an interpretable realization of the considered deadline- and energy-aware caching problem, its decisions still rely on predefined rules, thresholds, and priorities. In a dynamic multi-UAV network, however, content demand, user mobility, connectivity, cooperative service conditions, and UAV energy states evolve simultaneously and interact with one another. Explicitly designing rules for all such interactions becomes increasingly difficult and limits the scalability and adaptability of a rule-based approach as the system complexity grows. Therefore, we next formulate the cache placement and replacement process as an MDP, allowing these dynamically coupled trade-offs to be learned from interactions with the network environment.

\subsection{MDP Formulation}
\label{sub:mdp}
The dynamic caching problem is formulated as an MDP at the cache-decision epochs. Although the underlying network evolves at every discrete simulation step, the caching agent acts at cache-decision epochs separated by intervals of duration $\Delta_c$. Let $\xi$ denote the cache-decision epoch index, with the corresponding simulation time given by $t_\xi=\xi\Delta_c$. The MDP is defined as
\begin{equation}
\label{eq:mdp_definition}
\makebox[\linewidth][c]{%
$\displaystyle
\mathcal{M} =
\left\langle
\mathcal{S},
\mathcal{A},
\mathcal{P},
r
\right\rangle
$%
}
\end{equation}
where $\mathcal{S}$, $\mathcal{A}$, $\mathcal{P}$, and $r$
denote the state space, action space, state-transition probability distribution, and reward function, respectively.

\subsubsection{State space}
\label{subsub:state}
At each cache-decision epoch, the local state of UAV $n$ consists of:
(i) its current cache contents, (ii) the demand conditions of its associated
users, (iii) its user-association and network connectivity conditions,
(iv) its current energy consumption state, and (v) its deadline pressure.

\begin{equation}
\label{eq:state_space}
\makebox[\linewidth][c]{%
$\displaystyle
\begin{gathered}
s_{n,\xi}^{\mathrm{local}} =
\left(
\mathbf{x}_{n,\xi},
\mathbf{q}_{n,\xi}^{\mathrm{dem}},
\mathbf{g}_{n,\xi}^{\mathrm{conn}},
E_{n,\xi}^{\mathrm{comm}},
\boldsymbol{\psi}_{n,\xi}^{\mathrm{dl}}
\right), n\in\mathcal{N},
\\[5pt]
s_{\xi}^{\mathrm{global}} =
\left(
s_{1,\xi}^{\mathrm{local}},
\ldots,
s_{N,\xi}^{\mathrm{local}}
\right)
\end{gathered}$%
}
\end{equation}
The vectors $\mathbf{x}_{n,\xi}$, $\mathbf{q}_{n,\xi}^{\mathrm{dem}}$,
and $\mathbf{g}_{n,\xi}^{\mathrm{conn}}$ represent the cache state, local
demand state, and user-association and connectivity state of UAV $n$,
respectively. The term $E_{n,\xi}^{\mathrm{comm}}$ represents its
communication energy state. The global state conceptually consists of the
collection of the local states of all UAVs. The global state space
$\mathcal{S}$ consists of all feasible global network states.

Deadline pressure represents the recent demand for deadline-sensitive
content at each UAV. It is obtained by weighting the latest requests
according to their content deadlines and aggregating them by content type.
\begin{equation}
\label{eq:deadline_pressure}
\makebox[\linewidth][c]{%
$\displaystyle
\begin{gathered}
\psi_{n,\kappa,\xi}^{\mathrm{dl}}
=
\frac{
\displaystyle\sum_{i:\tau_i=\kappa}
\nu_{n,i,\xi}\omega_i^{\mathrm{urg}}
}{
\displaystyle
\max\left\{
\epsilon_{\psi},
\sum_{\kappa'\in\mathcal{T}}
\sum_{j:\tau_j=\kappa'}
\nu_{n,j,\xi}\omega_j^{\mathrm{urg}}
\right\}
},
\\[3pt]
\boldsymbol{\psi}_{n,\xi}^{\mathrm{dl}}
=
\left[
\psi_{n,\kappa,\xi}^{\mathrm{dl}}
\right]_{\kappa\in\mathcal{T}}
\end{gathered}$%
}
\end{equation}
Here, $\nu_{n,i,\xi}$ denotes the number of requests for content item $i$
arriving at UAV $n$ during the most recent cache-decision interval,
$\omega_i^{\mathrm{urg}}$ denotes the deadline urgency of content item $i$,
where a shorter delivery deadline corresponds to a larger urgency value,
$\kappa\in\mathcal{T}$ denotes a content type, and $\epsilon_{\psi}$ is a
small positive constant.

\subsubsection{Action space}

At each cache-decision epoch, the centralized agent determines the content items
to be removed from and added to each UAV cache. The joint action is defined as
\begin{equation}
\label{eq:action_space}
\makebox[\linewidth][c]{%
$\displaystyle
a_\xi
=
\left\{
\left(
\mathcal{C}_{n,\xi}^{-},
\mathcal{C}_{n,\xi}^{+}
\right)
\right\}_{n\in\mathcal{N}}
$%
}
\end{equation}
where $\mathcal{C}_{n,\xi}^{-}$ and
$\mathcal{C}_{n,\xi}^{+}$ denote, respectively, the sets of content items
removed from and added to UAV $n$. The collection of all feasible joint cache-update actions constitutes
the action space $\mathcal{A}$.

The action space also includes a no-operation action. For UAV $n$, the
no-operation action is defined as
\begin{equation}
\label{eq:no_op_action}
\makebox[\linewidth][c]{%
$\displaystyle
\mathcal{C}_{n,\xi}^{-}
=
\mathcal{C}_{n,\xi}^{+}
=
\varnothing
$%
}
\end{equation}
A cache-update action is considered feasible only if it satisfies the constraints in Section~\ref{subsub:cacheconstraints}. The environment evaluates these constraints when decoding the sampled action into an executable cache-update command.

\subsubsection{State--transition probability distribution}
The state-transition distribution describes the evolution of the global
network state over one cache-decision interval after applying a joint
cache-update action. It is defined as
\begin{equation}
\label{eq:transition_probability}
\makebox[\linewidth][c]{%
$\displaystyle
s_{\xi+1}^{\mathrm{global}}
\sim
\mathcal{P}\!\left(
\cdot
\mid
s_{\xi}^{\mathrm{global}},
a_{\xi}
\right)
$%
}
\end{equation}
Given the global state $s_{\xi}^{\mathrm{global}}$ and the selected joint
action $a_{\xi}$, the network evolves over the following cache-decision
interval according to the transition distribution $\mathcal{P}$. This
evolution accounts for cache-update execution, user mobility, request
generation, wireless resource variations, service outcomes, and energy
consumption. The resulting state
$s_{\xi+1}^{\mathrm{global}}$ summarizes the network conditions at the
next cache-decision epoch.

\subsubsection{Reward function}
To optimize the multi-objective caching problem using PPO, we construct
a scalarized interval-level reward that combines the three objectives. Since the caching agent acts at cache-decision epochs separated by intervals of duration
$\Delta_c$, the reward is computed at the end of each cache-decision interval. The interval reward integrates the service
performance and energy costs resulting from the selected cache-update
action. To jointly account for deadline misses, communication energy,
and inter-UAV energy imbalance, we define the interval-level reward as
\begin{equation}
\label{eq:interval_reward}
\makebox[\linewidth][c]{%
$\displaystyle
\begin{gathered}
r_\xi =
-\Biggl[
\frac{E_{\xi}^{\mathrm{total}}}{E_{\mathrm{ref}}}
+
\frac{\mathcal{I}_{E,\xi}}{\mathcal{I}_{E,\mathrm{ref}}}
+
\mathrm{DMR}_\xi
+
\mathcal{L}_{\mathrm{SFR},\xi}
\Biggr],
\\[5pt]
\mathcal{L}_{\mathrm{SFR},\xi}
=
\left[
\max\left(
0,\,
\frac{\mathrm{SFR}_{\xi}-\epsilon_{\mathrm{srv}}}
{1-\epsilon_{\mathrm{srv}}}
\right)
\right]^2
\end{gathered}
$%
}
\end{equation}
where $\mathrm{DMR}_{\xi}$, $\mathrm{SFR}_{\xi}$,
$E_{\xi}^{\mathrm{total}}$, and $\mathcal{I}_{E,\xi}$ denote the
deadline-miss rate, service failure ratio (SFR), total communication energy,
and inter-UAV energy imbalance during interval $\xi$, respectively.
Specifically, $\mathrm{SFR}_{\xi}$, i.e., the dropped-request ratio, is the
fraction of requests generated during interval $\xi$ that cannot be served
owing to the absence of a feasible delivery path. The normalization constants $E_{\mathrm{ref}}$ and $\mathcal{I}_{E,\mathrm{ref}}$ are set to the 95th percentiles of the
corresponding quantities observed in offline calibration rollouts sampled
from the training-scenario distribution. They are fixed before PPO training
and are not updated using evaluation episodes. The term $\mathcal{L}_{\mathrm{SFR},\xi}$ penalizes service failure ratios above the admissible threshold $\epsilon_{\mathrm{srv}}$. The SFR penalty prevents the policy from artificially reducing the DMR by allowing excessive request drops.

The above MDP reveals two key requirements for the caching policy. First, the state variables describing demand, connectivity, deadline pressure, and communication-energy conditions evolve over time, making recent observation histories informative for the current cache decision. Second, cooperative caching and inter-UAV energy balancing require each local condition to be interpreted in the context of the entire UAV network. Accordingly, GRU-DEPPO employs UAV-specific and network-wide temporal representations and uses them within a centralized policy for joint cache-update decision making.

\subsection{GRU-based Temporal Representation}
\label{sub:gru_encoding}

The MDP state contains time-varying information on local content demand, connectivity, cache conditions, deadline pressure, and UAV communication energy. Since these observations are temporally correlated, relying only on the current state may not sufficiently characterize the evolving local and network-wide conditions relevant to cache placement and replacement.
Moreover, directly providing the temporal histories of all UAVs to the centralized controller would increase the input dimension and communication overhead. We therefore employ local and global GRUs to construct compact temporal representations at the UAV and network levels, respectively.

For each UAV $n$, a shared local GRU independently encodes its most recent $L$ local observations. Let $\mathbf{o}_{n,\xi}^{\mathrm{loc}}$ denote the local observation vector at cache-decision epoch $\xi$. The UAV-specific temporal representation is obtained as
\begin{equation}
\label{eq:local_gru_encoding}
\mathbf{z}_{n,\xi}
=
f_{\theta_{\mathrm{loc}}}
\left(
\mathbf{o}_{n,\xi-L+1}^{\mathrm{loc}},
\ldots,
\mathbf{o}_{n,\xi}^{\mathrm{loc}}
\right),
\qquad n\in\mathcal{N}.
\end{equation}
The GRU parameters are shared across UAVs, while a separate representation $\mathbf{z}_{n,\xi}$ is generated for each UAV, thereby preserving UAV-specific temporal characteristics. The detailed composition of $\mathbf{o}_{n,\xi}^{\mathrm{loc}}$ is provided in Appendix~\ref{app:gru_input}.

In parallel, the MBS maintains a recent history of compact network-wide context vectors constructed from information available at the centralized controller. These contexts summarize network activity, service quality, service modes, communication energy, demand variation, and inter-UAV energy imbalance. The resulting context tokens are processed by the
global GRU as
\begin{equation}
\label{eq:global_gru_representation}
\mathbf{z}_{\xi}^{\mathrm{global}}
=
f_{\theta_{\mathrm{global}}}
\left(
\mathbf{e}_{\xi-L+1}^{\mathrm{net}},
\ldots,
\mathbf{e}_{\xi}^{\mathrm{net}}
\right),
\end{equation}
where $\mathbf{e}_{\xi}^{\mathrm{net}}$ denotes the embedded network-wide context at epoch $\xi$. Its detailed composition and embedding procedure are described in Appendix~\ref{app:gru_input}.

Before PPO training, the shared local GRU is pretrained using local-observation trajectories collected with a lightweight caching policy. Let $\mathcal{D}_{\mathrm{pre}}$ denote the resulting pretraining dataset. Each observation sequence is associated with semantic summary targets derived from its observed demand, cache, deadline, service, and energy conditions. A lightweight prediction head is trained together with the local GRU using
\begin{equation}
\label{eq:gru_pretraining}
\mathcal{L}_{\mathrm{pre}}
=
\operatorname{SmoothL1}
\left(
\widehat{\mathbf{y}}_{n,\xi}^{\mathrm{base}},
\mathbf{y}_{n,\xi}^{\mathrm{base}}
\right)
+
\lambda_{\mathrm{type}}
\operatorname{MSE}
\left(
\widehat{\mathbf{y}}_{n,\xi}^{\mathrm{type}},
\mathbf{y}_{n,\xi}^{\mathrm{type}}
\right),
\end{equation}
where the base component summarizes the local network conditions and the type component represents the request-type distribution. After pretraining, the prediction head is discarded and the shared local GRU is kept fixed during PPO training. The global GRU, in contrast, is trained jointly with the centralized policy. At each cache-decision epoch, only the compact local representation $\mathbf{z}_{n,\xi}$ is provided by each UAV to the MBS.

\subsection{GRU-assisted PPO optimization}

At each cache-decision epoch, the recent local observation history of each UAV is encoded by the shared local GRU, while the MBS encodes the recent network-wide context using the global GRU. The resulting local and global temporal representations are fused and provided to a centralized PPO actor--critic~\citep{PPO}. The actor selects feasible cache placement and replacement actions, and the resulting service, deadline, and communication-energy outcomes are used to update the policy.

\subsubsection{PPO input}

The MBS constructs the PPO input by concatenating the local temporal representations of all UAVs with the global representation:
\begin{equation}
\label{eq:fused_representation}
\makebox[\linewidth][c]{%
$\displaystyle
\mathbf{h}_{\xi}
=
\left[
\mathbf{z}_{1,\xi};
\ldots;
\mathbf{z}_{N,\xi};
\mathbf{z}_{\xi}^{\mathrm{global}}
\right]
$%
}
\end{equation}
where $\mathbf{h}_{\xi}$ represents the temporal network context used for cache-update decision making at epoch $\xi$.

\subsubsection{Factorized cache-update action}

To represent cache placement and replacement decisions efficiently, the joint cache-update action is factorized across UAVs as
\begin{equation}
\label{eq:factorized_action}
\makebox[\linewidth][c]{%
$\displaystyle
a_{\xi}
=
\left[
\mathbf{a}_{1,\xi};
\ldots;
\mathbf{a}_{N,\xi}
\right],
\qquad
\mathbf{a}_{n,\xi}
=
\left(
o_{n,\xi},
b_{n,\xi},
i_{n,\xi}
\right)
$%
}
\end{equation}
where $o_{n,\xi}$ specifies the cache operation, $b_{n,\xi}$ identifies the corresponding cache slot, and $i_{n,\xi}$ identifies the selected content item. The operation set includes a no-operation choice, in which the slot and content components are ignored.

Let $\mathbf{m}_{\xi}^{\mathrm{mask}}$ denote the feasibility mask constructed from the current cache state and network conditions. The masked policy is factorized as
\begin{equation}
\label{eq:factorized_policy}
\makebox[\linewidth][c]{%
$\displaystyle
\begin{aligned}
\pi_{\theta_{\pi}}^{\mathrm{mask}}
\left(
a_{\xi}
\mid
\mathbf{h}_{\xi},
\mathbf{m}_{\xi}^{\mathrm{mask}}
\right)
=
\prod_{n\in\mathcal{N}}
&
\pi_{\theta_{\pi}}^{o}
\left(
o_{n,\xi}
\mid
\mathbf{h}_{\xi},
\mathbf{m}_{\xi}^{\mathrm{mask}}
\right)
\\[-1pt]
{}\times&
\pi_{\theta_{\pi}}^{b}
\left(
b_{n,\xi}
\mid
\mathbf{h}_{\xi},
\mathbf{m}_{\xi}^{\mathrm{mask}}
\right)
\\[-1pt]
{}\times&
\pi_{\theta_{\pi}}^{i}
\left(
i_{n,\xi}
\mid
\mathbf{h}_{\xi},
\mathbf{m}_{\xi}^{\mathrm{mask}}
\right).
\end{aligned}
$%
}
\end{equation}

The mask excludes infeasible component choices before sampling. The sampled action is decoded into the remove/add sets $\left(\mathcal{C}_{n,\xi}^{-},\mathcal{C}_{n,\xi}^{+}\right)$, and the resulting cache update is executed subject to the cache-capacity and cache-update feasibility constraints defined in Section~\ref{subsub:cacheconstraints}.

\subsubsection{PPO-based policy optimization}

After applying the selected cache-update action, the network evolves for one cache-decision interval $\Delta_c$. The resulting request service outcomes, deadline misses, service failures, and communication energy consumption determine the interval-level reward $r_{\xi}$ defined in Section~\ref{sub:mdp}.

The actor and critic are trained using standard PPO with generalized advantage estimation (GAE)~\citep{PPO}. Let $\widehat{A}_{\xi}$ denote the GAE advantage estimate and let $\rho_{\xi}$ denote the probability ratio between the updated policy and the policy used during rollout collection. The clipped policy objective is
\begin{equation}
\label{eq:ppo_clipped_objective}
\makebox[\linewidth][c]{%
$\displaystyle
\mathcal{L}_{\mathrm{clip}}
=
\mathbb{E}_{\xi}
\left[
\min
\left(
\rho_{\xi}\widehat{A}_{\xi},
\operatorname{clip}
\left(
\rho_{\xi},
1-\epsilon_{\mathrm{clip}},
1+\epsilon_{\mathrm{clip}}
\right)
\widehat{A}_{\xi}
\right)
\right].
$%
}
\end{equation}

The critic is trained using the corresponding return estimates, and entropy regularization is applied to encourage policy exploration. The PPO training parameters are summarized in the Appendix.
The complete GRU-DEPPO training procedure is summarized in
Algorithm~\ref{alg:gru_deppo}.

\begin{algorithm}[t]
\caption{GRU-DEPPO Training}
\label{alg:gru_deppo}
\begin{algorithmic}[1]

\Require $\mathcal{D}_{\mathrm{pre}}$, $N_{\mathrm{ep}}$, $B_{\mathrm{roll}}$
\Ensure Trained GRU-DEPPO policy

\State Train $f_{\theta_{\mathrm{loc}}}$ using $\mathcal{L}_{\mathrm{pre}}$
\State Fix $\theta_{\mathrm{loc}}$ and initialize
$f_{\theta_{\mathrm{global}}}$, PPO, and $\mathcal{B}$

\For{$e=1,\ldots,N_{\mathrm{ep}}$}
    \State Reset environment and histories

    \While{episode is not terminated}

        \Statex \textbf{Stage 1: Temporal encoding}
        \For{$n\in\mathcal{N}$}
            \State $\mathbf{z}_{n,\xi}
            \gets f_{\theta_{\mathrm{loc}}}
            (\mathbf{o}_{n,\xi-L+1}^{\mathrm{loc}},\ldots,
            \mathbf{o}_{n,\xi}^{\mathrm{loc}})$
        \EndFor
        \State $\mathbf{z}_{\xi}^{\mathrm{global}}
        \gets f_{\theta_{\mathrm{global}}}
        (\mathbf{e}_{\xi-L+1}^{\mathrm{net}},\ldots,
        \mathbf{e}_{\xi}^{\mathrm{net}})$
        \State Construct $\mathbf{h}_{\xi}$

        \Statex \textbf{Stage 2: Cache-action generation}
        \State Construct $\mathbf{m}_{\xi}^{\mathrm{mask}}$
        \State Sample
        $a_{\xi}\sim
        \pi_{\theta_{\pi}}^{\mathrm{mask}}
        (\cdot\mid\mathbf{h}_{\xi},
        \mathbf{m}_{\xi}^{\mathrm{mask}})$
        \State Decode $a_{\xi}$ into
        $\{(\mathcal{C}_{n,\xi}^{-},
        \mathcal{C}_{n,\xi}^{+})\}_{n\in\mathcal{N}}$

        \Statex \textbf{Stage 3: Environment interaction}
        \State Apply $a_{\xi}$ and evolve for $\Delta_c$
        \State Observe $s_{\xi+1}^{\mathrm{global}}$ and $r_{\xi}$
        \State Store transition in $\mathcal{B}$

        \If{$|\mathcal{B}|\geq B_{\mathrm{roll}}$}
            \Statex \textbf{Stage 4: PPO optimization}
            \State Compute GAE and return estimates
            \State Update $f_{\theta_{\mathrm{global}}}$ and PPO
            \State $\mathcal{B}\gets\emptyset$
        \EndIf

    \EndWhile
\EndFor

\end{algorithmic}
\end{algorithm}

\section{Results and discussion}
\label{sec:results}

This section evaluates GRU-DEPPO from multiple perspectives. We first compare its overall performance with the baseline schemes in terms of deadline-miss rate, content-type-specific deadline-miss performance, total communication energy consumption, and inter-UAV energy imbalance. We then examine its robustness under more stringent content-specific deadline requirements without retraining. Additional analyses investigate cache-update energy, request service modes, and adaptation to abrupt changes in content demand. Finally, we conduct an ablation study to quantify the contribution of the GRU-based temporal representation to both learning behavior and caching performance.

We compare GRU-DEPPO with LRU as a conventional cache replacement baseline, DE-UBAC (Algorithm~\ref{alg:de_ubac}) as a rule-based caching scheme, and a TLPPO-based caching baseline adapted from the caching component of TLPPO~\citep{DRL4}. The TLPPO-based baseline employs PPO to determine content placement and replacement actions based on recently observed local content demand and the current UAV cache state. Its policy is trained to maximize interval-level cache-hit performance. All reported performance evaluation results are averaged over 30 independent random seeds. The detailed simulation parameters are listed in Table~\ref{tab:simulation_parameters}.

\subsection{Simulation setup}
\begin{table}[t]
\caption{Main simulation parameters.}
\centering
\renewcommand{\arraystretch}{1.05}
\setlength{\tabcolsep}{5pt}
\footnotesize
\begin{tabularx}{\columnwidth}{
@{} >{\centering\arraybackslash}X
>{\centering\arraybackslash}X @{}
}
\hline
Parameter & Value \\
\hline

$N$ & 10 \\
$|\mathcal{U}|$ & 150 \\
$\mathcal{W}$ & $3000~\mathrm{m}\times3000~\mathrm{m}$ \\
UAV height & $100~\mathrm{m}$ \\
User speed & $[1,20]~\mathrm{m/s}$ \\
User groups & 3 \\
$R_g$ & $100~\mathrm{m}$ \\
$\alpha_u$
& $[0.4]/[0.5]/[0.8]$ \\

$|\mathcal{C}|,|\mathcal{T}|$ & $200,\,3$ \\
$S$ & $25~\mathrm{Mbit}$ \\
$K_n S$ & $200~\mathrm{Mbit}$ \\
$\{\delta_{\kappa}\}_{\kappa\in\mathcal{T}}$
& $[2.5,4.0,5.5]~\mathrm{s}$ \\

$\alpha$ & 0.6 \\
$\lambda_{u,t}$
& $0.6/0.7/0.8$ \\
$\operatorname{std}(\epsilon_{i,g,t})$ & 0.02 \\
$\beta_{\mathrm{shift}}$ & 0.4 \\

Simulation duration & $1000~\mathrm{s}$ \\
Simulation time step & $1~\mathrm{s}$ \\
$\Delta_c$ & $10~\mathrm{s}$ \\

$f_u,f_{\mathrm{UAV}},f_{\mathrm{MBS}}$
& $5,\,4.5,\,4~\mathrm{GHz}$ \\

Link bandwidths
& $40/30/10~\mathrm{MHz}$ \\

$P_n^{\mathrm{tx}}$ & $30~\mathrm{dBm}$ \\
MBS transmit power & $43~\mathrm{dBm}$ \\
UAV receive power & $1.2~\mathrm{W}$ \\

$X,Y$ & $11.9,\,0.13$ \\
$\eta_{\mathrm{LoS}},\eta_{\mathrm{NLoS}}$
& $1.6,\,23.0~\mathrm{dB}$ \\
$N_0$ & $3.98\times10^{-21}~\mathrm{W/Hz}$ \\
$\Gamma_{\min}$ & $-10~\mathrm{dB}$ \\
$R_{\min}$ & $10^{5}~\mathrm{bps}$ \\

DE-UBAC: $\lambda_{\mathrm{dl}},\lambda_{\mathrm{per}},
\tau_H,\lambda_{\mathrm{up}}$
& $0.5,\,0.5,\,0.1,\,0.25$ \\
\hline
\end{tabularx}
\vspace{1mm}
\parbox{\columnwidth}{
\scriptsize
Group-specific parameters are listed in the order G1, G2, and G3.

For grouped link parameters, values are reported in the order
User--UAV, UAV--UAV, and MBS--UAV.
}
\label{tab:simulation_parameters}
\end{table}
We consider a multi-UAV service area in which users are divided into multiple groups with distinct mobility and content-demand characteristics.

\subsubsection{User mobility and spatial locality}
To model heterogeneous local conditions, users follow a group hotspot-aware mobility model. Users are divided into
several behavior groups, and each group is associated with a spatial hotspot representing its preferred area of activity. Users within the same group share the same hotspot region, while their
movement trajectories are generated independently. Each user is also
assigned an individual hotspot attraction probability to represent
different degrees of spatial locality.

Let $g(u)$ denote the group of user $u$, and let $\mathbf{h}_{g(u)}$ and $R_{g(u)}$ denote the hotspot center and radius of that group, respectively. At each waypoint selection, user $u$ selects a destination around its group hotspot with probability $\alpha_u$; otherwise, the destination is selected according to the ordinary random-waypoint process~\citep{RWP}:
\begin{equation}
\label{eq:mobility_waypoint}
\makebox[\linewidth][c]{%
$\displaystyle
\mathbf{w}_{u,t}
\sim
\begin{cases}
\mathcal{D}\!\left(\mathbf{h}_{g(u)},R_{g(u)}\right),
& \text{with probability } \alpha_u,\\[3pt]
\mathcal{D}_{\mathrm{RWP}}(\mathcal{W}),
& \text{with probability } 1-\alpha_u
\end{cases}
$%
}
\end{equation}
where $\mathcal{D}(\mathbf{h}_{g(u)},R_{g(u)})$ denotes the waypoint
distribution within the hotspot region centered at $\mathbf{h}_{g(u)}$ with radius $R_{g(u)}$, and $\mathcal{W}$ denotes the service area. The attraction probability $\alpha_u$ is sampled according to the user's group, thereby allowing users in different groups to exhibit different degrees of spatial locality.

As users move between spatial regions, their serving UAVs and local network conditions change over time. Consequently, different UAVs
observe different user populations, request compositions, connectivity conditions, and service loads. In addition, the hotspot locations are shifted during the episode to represent a change in the spatial mobility regime.

Spatial locality is accompanied by group-conditioned request behavior. Users belonging to the same group share a common time-varying local content distribution $p_{i,g(u),t}$, whereas different groups exhibit different content preferences. The request-generation probability may also vary across users or groups. Consequently, as users move across the service area and their serving UAVs change, different UAVs observe distinct local traffic patterns resulting from different mixtures of user groups. At the same change point, the group-conditioned content-demand distributions are also shifted to represent temporal changes in user preferences.

\subsection{Simulation results}

\begin{figure*}[t]
\centering
\begin{subfigure}[t]{0.22\textwidth}
\centering
\includegraphics[width=\linewidth]{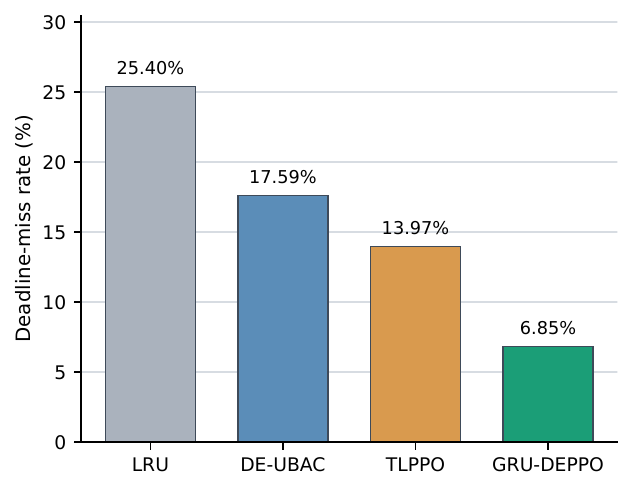}
\caption{Deadline-miss rate}
\label{fig:deadline_miss_rate}
\end{subfigure}
\hfill
\begin{subfigure}[t]{0.30\textwidth}
\centering
\includegraphics[width=\linewidth]{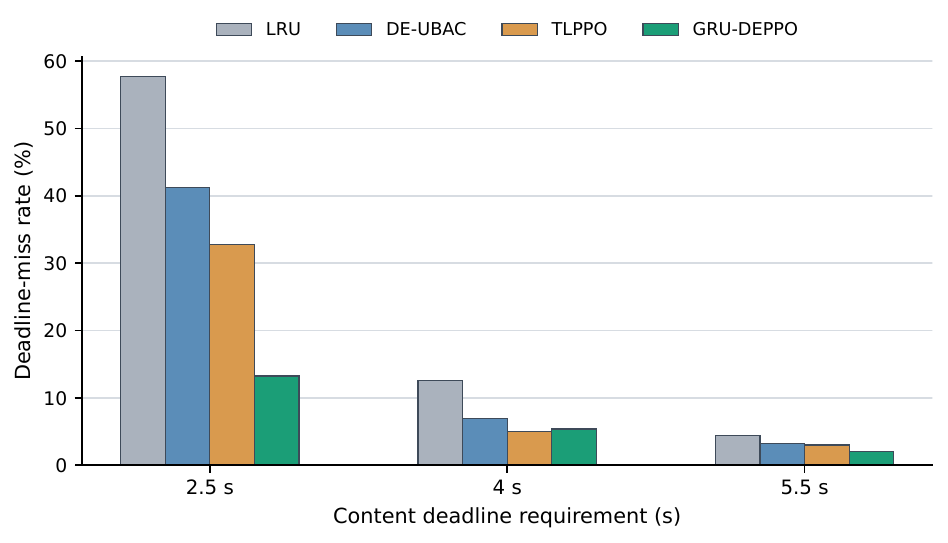}
\caption{DMR by content type}
\label{fig:content_type_dmr_default}
\end{subfigure}
\hfill
\begin{subfigure}[t]{0.22\textwidth}
\centering
\includegraphics[width=\linewidth]{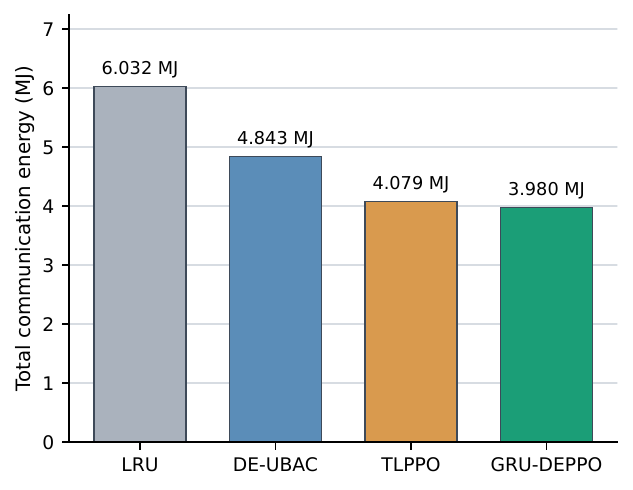}
\caption{Total communication energy}
\label{fig:total_communication_energy}
\end{subfigure}
\hfill
\begin{subfigure}[t]{0.22\textwidth}
\centering
\includegraphics[width=\linewidth]{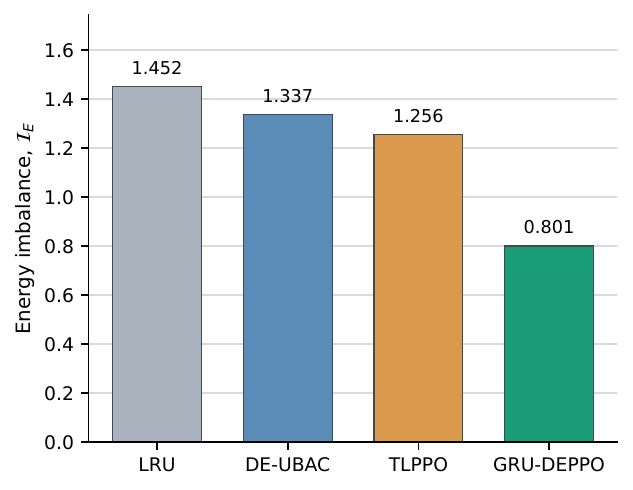}
\caption{Inter-UAV energy imbalance}
\label{fig:energy_imbalance}
\end{subfigure}

\caption{Performance with deadlines of $[2.5,4.0,5.5]$~s}
\label{fig:overall_performance}
\end{figure*}

\begin{figure*}[t]
\centering

\begin{subfigure}[t]{0.22\textwidth}
\centering
\includegraphics[width=\linewidth]{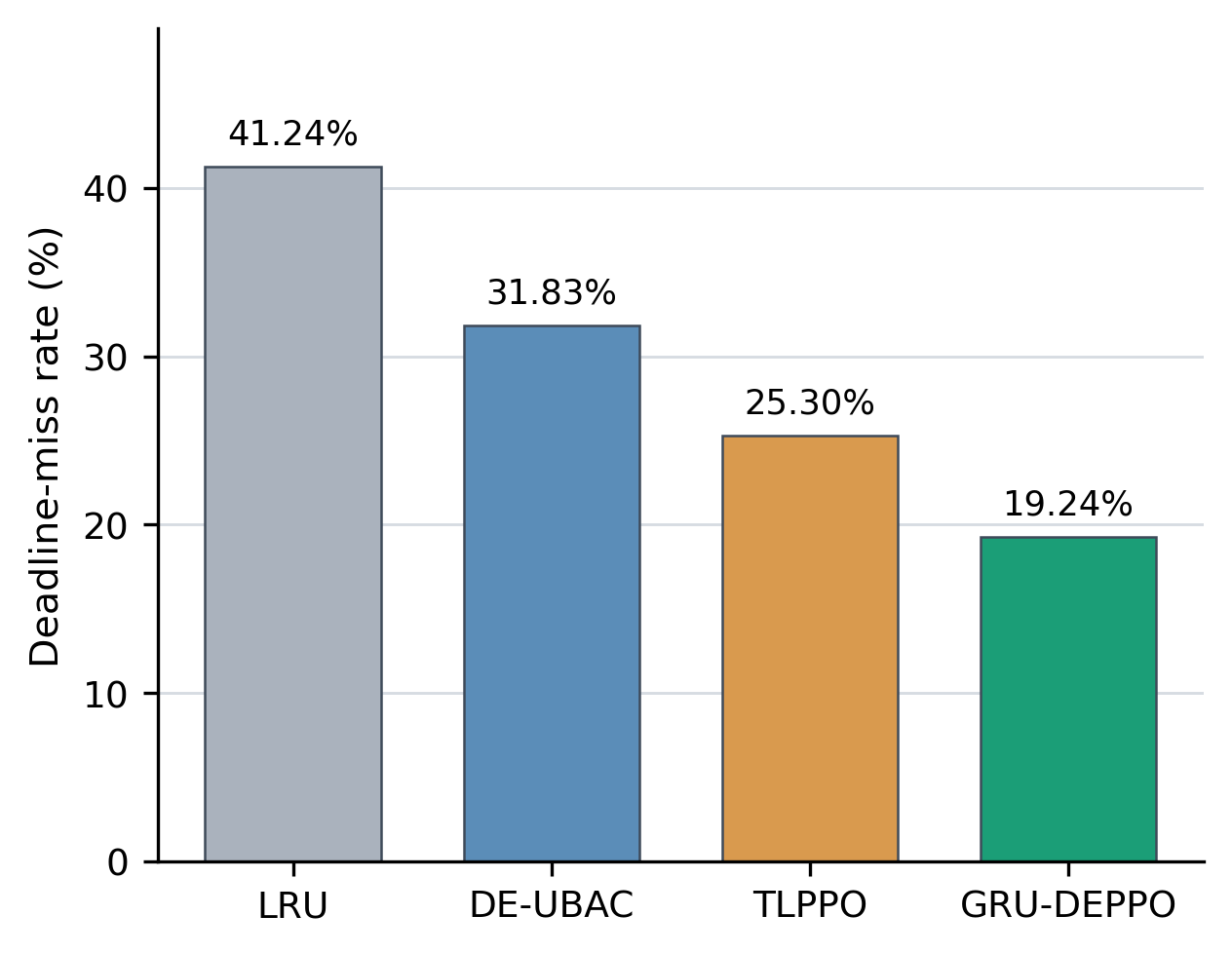}
\caption{Deadline-miss rate}
\label{fig:deadline_miss_rate2}
\end{subfigure}
\hfill
\begin{subfigure}[t]{0.30\textwidth}
\centering
\includegraphics[width=\linewidth,trim=0 2mm 0 0,clip]{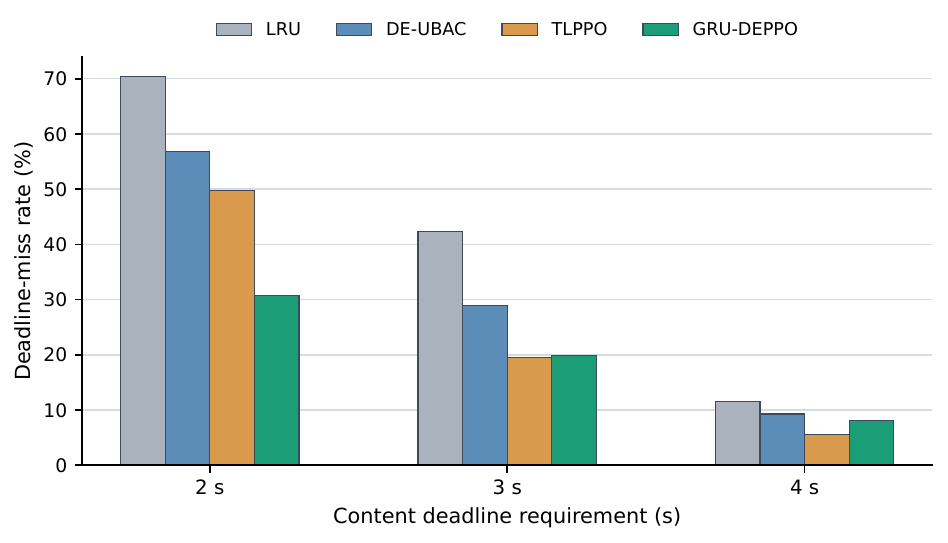}
\caption{DMR by content type}
\label{fig:content_type_dmr_strict}
\end{subfigure}
\hfill
\begin{subfigure}[t]{0.22\textwidth}
\centering
\includegraphics[width=\linewidth]{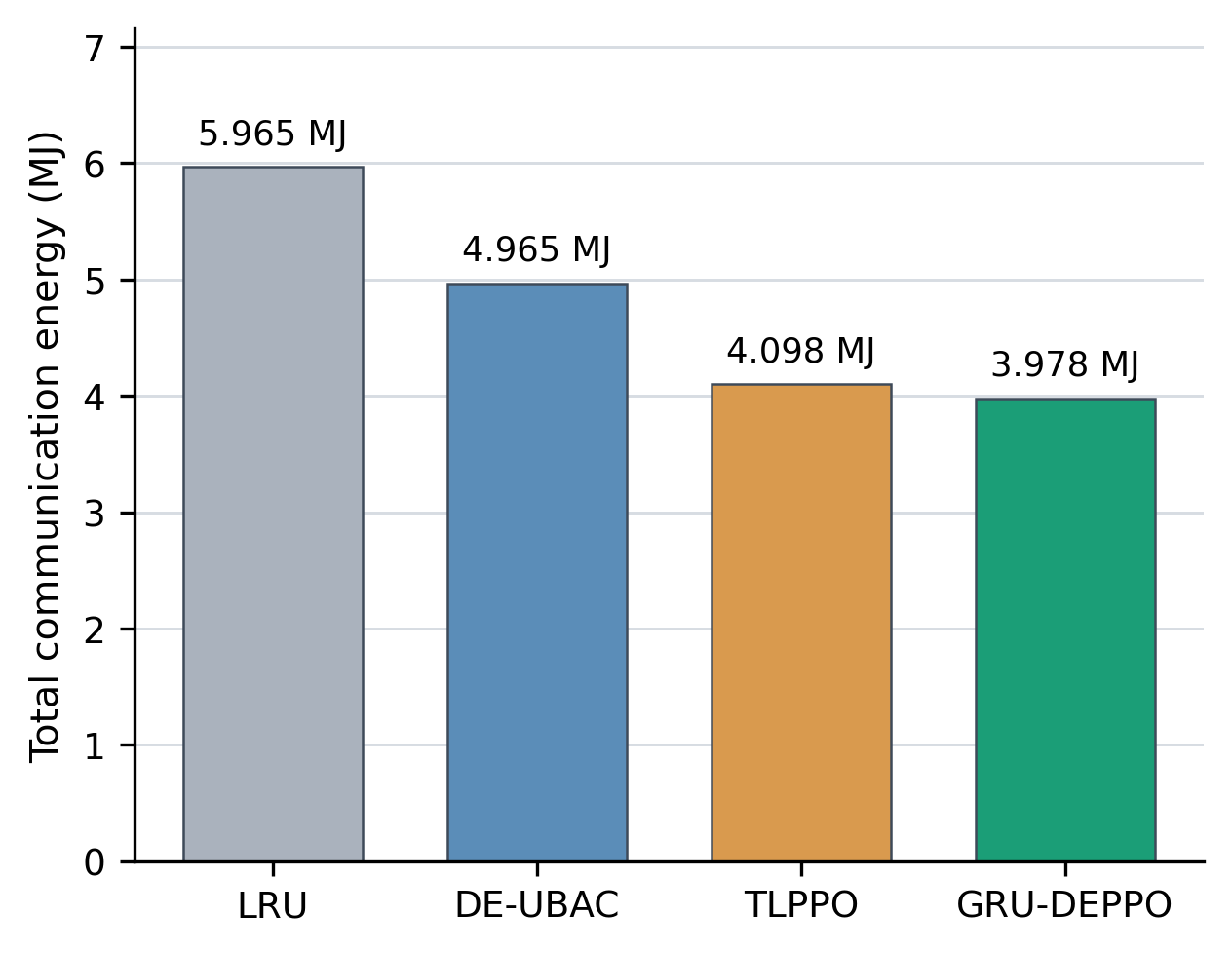}
\caption{Total communication energy}
\label{fig:total_communication_energy2}
\end{subfigure}
\hfill
\begin{subfigure}[t]{0.22\textwidth}
\centering
\includegraphics[width=\linewidth]{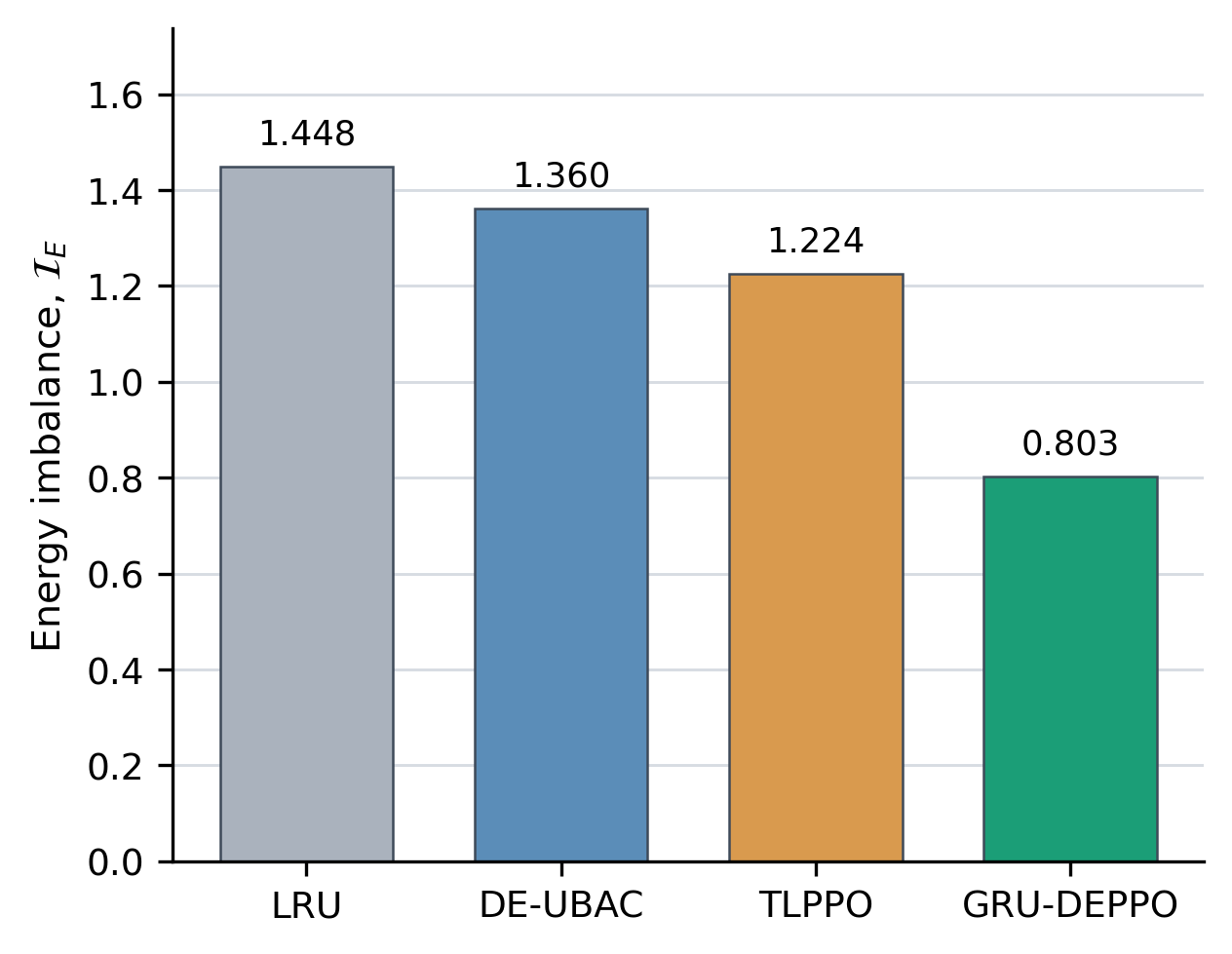}
\caption{Inter-UAV energy imbalance}
\label{fig:energy_imbalance2}
\end{subfigure}

\caption{Performance with deadlines of $[2.0,3.0,4.0]$~s}
\label{fig:strict_performance}
\end{figure*}

Fig.~\ref{fig:overall_performance} compares the overall performance of the considered caching schemes in terms of deadline-miss rate, content-type-specific deadline-miss performance, total communication energy consumption, and inter-UAV energy imbalance.

As shown in Fig.~\ref{fig:deadline_miss_rate}, LRU exhibits the highest deadline-miss rate because its cache replacement decisions are based solely on content recency and do not explicitly account for local demand conditions or delivery deadlines. DE-UBAC improves upon LRU by incorporating localized user behavior, deadline misses, cooperative delivery feasibility, and UAV energy conditions into its hierarchical cache-decision rules. However, these factors are coordinated through predefined priorities and thresholds, which limits the flexibility of the rule-based policy under rapidly changing network conditions.

TLPPO achieves a substantially lower deadline-miss rate than LRU and DE-UBAC. By observing recent local content demand and the current cache state, its PPO-based policy can proactively adjust content placement rather than reacting only after individual cache misses. Moreover, because the policy is trained to maximize interval-level cache-hit performance, it substantially reduces the average delivery delay. Nevertheless, GRU-DEPPO achieves the lowest deadline-miss rate among all compared schemes, reducing the DMR by approximately 51\% relative to TLPPO. This improvement can be attributed to two key differences from TLPPO. First, GRU-DEPPO exploits local and global GRUs to capture the temporal evolution of spatially heterogeneous local network conditions. Second, its decision model explicitly accounts for content-specific deadlines, rather than focusing primarily on interval-level cache-hit performance.

Fig.~\ref{fig:content_type_dmr_default} further shows the DMR for each content type. GRU-DEPPO achieves a substantial improvement for contents with the most stringent 2.5-s deadline, whereas TLPPO achieves a slightly lower DMR for contents with the 4.0-s deadline. For contents with the relatively relaxed 5.5-s deadline, GRU-DEPPO again achieves the lowest DMR. These results indicate that GRU-DEPPO does not improve deadline-miss performance uniformly across all content types. Instead, the deadline-dependent performance suggests that the policy distinguishes heterogeneous content-specific delivery requirements and places greater emphasis on satisfying requests with tighter deadlines. In particular, the large reduction observed for the most stringent content type contributes substantially to the lower overall DMR of GRU-DEPPO.

Interestingly, TLPPO achieves a slightly lower average delivery delay than GRU-DEPPO, with approximately 1.99~s and 2.01~s, respectively. However, their corresponding deadline-miss rates are approximately 13.97\% and 6.85\%, respectively. This contrast indicates that a lower average delivery delay does not necessarily lead to better deadline satisfaction under heterogeneous content-specific deadlines. The service failure ratios remain within a relatively narrow range, with 11.66\% for GRU-DEPPO, 10.59\% for LRU, 10.62\% for DE-UBAC, and
10.69\% for TLPPO. Although GRU-DEPPO exhibits a slightly higher SFR, the approximately one-percentage-point difference indicates that its substantial reduction in deadline-miss rate is not attributable to excessive request dropping.

Fig.~\ref{fig:total_communication_energy} shows the total communication energy consumed by all UAVs throughout the simulation period. LRU incurs the highest energy consumption among the compared schemes, whereas GRU-DEPPO achieves the lowest total communication energy.
The improvement over TLPPO is relatively modest compared with those over LRU and DE-UBAC. This is because TLPPO also promotes edge cache hits by maximizing interval-level cache-hit performance. Consequently, a large fraction of requests can be served through local or cooperative UAV caching rather than relying on MBS-assisted delivery, which naturally reduces communication energy consumption. Although the reduction in total communication energy relative to TLPPO is relatively modest, a substantially larger difference is observed in the inter-UAV energy imbalance, as shown in Fig.~\ref{fig:energy_imbalance}. GRU-DEPPO achieves an energy-imbalance value of 0.801, compared with 1.256 for TLPPO, corresponding to an
approximately 36\% reduction.

This difference arises because TLPPO primarily improves network-wide cache-hit performance and therefore can reduce the overall communication cost, but it does not explicitly regulate how communication energy is distributed among individual UAVs. In contrast, GRU-DEPPO explicitly incorporates inter-UAV energy imbalance into its decision objective, discouraging cache and service decisions that repeatedly concentrate communication load on a subset of UAVs. In particular, among the MBS-direct UAVs that also serve as relay gateways for edge UAVs without direct MBS connectivity, GRU-DEPPO achieves an energy-consumption CV of 0.568, compared with 0.977 for LRU, 0.908 for DE-UBAC, and 0.922 for TLPPO. This result indicates that the overall reduction in inter-UAV energy imbalance is especially pronounced among MBS-direct UAVs, where relay-related communication burden is more likely to be concentrated.

Overall, GRU-DEPPO effectively captures the temporal evolution of spatially heterogeneous local network conditions through its GRU-based representations. Moreover, by explicitly incorporating content-specific deadlines and inter-UAV energy imbalance into the PPO-based decision framework, rather than focusing solely on average delivery delay or cache-hit performance, GRU-DEPPO achieves a better overall trade-off among deadline satisfaction, communication energy consumption, and inter-UAV energy imbalance.

Fig.~\ref{fig:strict_performance} further evaluates the compared schemes under more stringent delivery requirements. Considering that the PPO-based caching schemes exhibit average delivery delays of approximately 2~s under the original setting, the content-type-specific deadlines are tightened from $[2.5,4.0,5.5]$~s to $[2.0,3.0,4.0]$~s. The PPO-based policies obtained under the original simulation setting
are evaluated without retraining under the tightened deadlines. An interesting observation can be made from Fig.~\ref{fig:deadline_miss_rate2}. Compared with the original deadline setting, the DMR of TLPPO increases by 11.33 percentage points, whereas that of GRU-DEPPO increases by 12.39 percentage points. This degradation is not necessarily attributable solely to the caching policy itself. When the delivery deadlines become more stringent, requests become more sensitive to transient resource limitations caused by, for example, temporary concentrations of users or request arrivals in particular network regions. Under such conditions, the available communication resources or feasible delivery paths may be insufficient to satisfy all requests within the shortened deadlines, regardless of the caching decision. Consequently, tighter deadline requirements can expose environment-intrinsic performance limitations that are difficult to fully mitigate through cache placement alone.

Fig.~\ref{fig:content_type_dmr_strict} further decomposes the DMR shown in Fig.~\ref{fig:deadline_miss_rate2} according to the content-specific deadline requirement. As expected, contents with the most stringent 2.0-s deadline exhibit the highest DMR across all compared schemes. However, GRU-DEPPO shows a different performance trade-off from TLPPO. Although GRU-DEPPO exhibits somewhat higher DMRs for contents with the relatively relaxed 3.0-s and 4.0-s deadlines, it achieves a substantial reduction in DMR for contents with the most stringent 2.0-s deadline. This result suggests that, as the delivery requirements become more stringent, the advantage of GRU-DEPPO becomes increasingly concentrated on the most deadline-sensitive contents. In other words, the policy places greater emphasis on satisfying requests with tighter deadline requirements rather than uniformly reducing the DMR of all content types.

Despite the slightly higher DMRs for the relatively relaxed content types, GRU-DEPPO achieves a 24.0\% lower overall DMR than TLPPO under the stringent deadline setting. This result indicates that, in the observed performance trade-off, the substantial reduction in deadline misses for the most deadline-sensitive content outweighs the smaller performance degradations for the relatively relaxed content types.
Moreover, as shown in Figs.~\ref{fig:total_communication_energy2} and~\ref{fig:energy_imbalance2}, GRU-DEPPO maintains its advantages in both total communication energy consumption and inter-UAV energy imbalance. These results indicate that GRU-DEPPO maintains a favorable overall performance trade-off even when the content-delivery requirements become substantially more stringent.

\begin{figure*}[pos=t]
\centering

\begin{subfigure}[t]{0.48\textwidth}
\centering
\includegraphics[width=\linewidth]{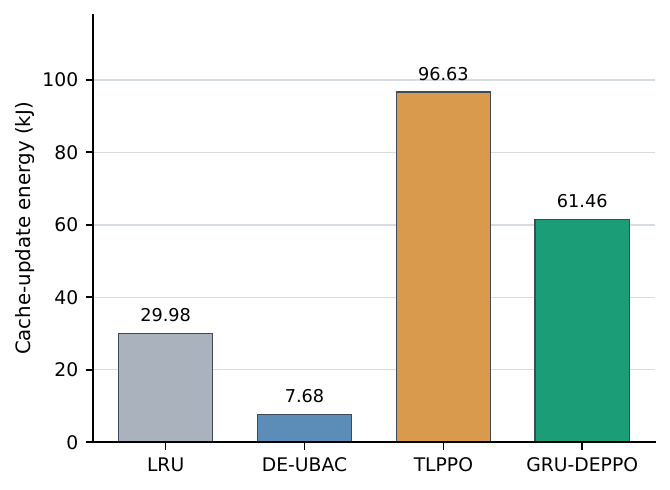}
\caption{Cache-update energy}
\label{fig:cache_update_energy}
\end{subfigure}
\hfill
\begin{subfigure}[t]{0.48\textwidth}
\centering
\includegraphics[width=\linewidth]{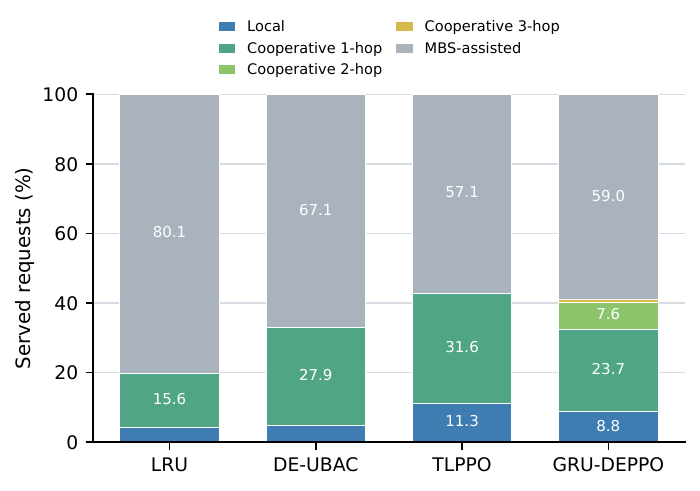}
\caption{Distribution of request service modes}
\label{fig:service_distribution}
\end{subfigure}

\caption{Energy and service characteristics}
\label{fig:service_distribution_energy}
\end{figure*}

We next analyze the composition of communication-energy consumption and the distribution of request service modes to further investigate how the compared caching schemes exploit the available service opportunities under dynamic network conditions. Fig.~\ref{fig:cache_update_energy} separately compares the communication
energy consumed for cache updates during the simulation period. LRU and
DE-UBAC incur relatively low cache-update energy among the compared
schemes, but for different reasons. LRU performs cache replacement using
a simple recency-based rule and therefore triggers relatively limited
cache reconfiguration. In contrast, DE-UBAC explicitly suppresses a
replacement when the expected priority gain is insufficient to justify
the corresponding update-energy cost, thereby avoiding low-benefit cache
updates. Nevertheless, its predefined decision rules provide less
flexibility in adapting the overall cache configuration than the
learning-based schemes. Consequently, low cache-update energy alone does
not necessarily imply lower total communication energy.

Among the RL-based schemes, TLPPO consumes approximately 36.4\% more cache-update energy than GRU-DEPPO. TLPPO is trained primarily to
maximize interval-level cache-hit performance and does not explicitly account for the communication cost associated with cache updates.
Consequently, the policy may still favor cache reconfiguration even when the expected gain in cache-hit performance is relatively small compared with the additional energy required for the update, leading to more aggressive cache-update behavior. In contrast, GRU-DEPPO explicitly incorporates total communication energy consumption into its decision objective. Therefore, the policy accounts for the energy cost incurred by cache reconfiguration as part of the overall communication-energy trade-off. This enables GRU-DEPPO to achieve lower cache-update energy than TLPPO while simultaneously attaining a lower deadline-miss rate and lower total communication energy consumption. Fig.~\ref{fig:service_distribution} presents the distribution of request service modes for each caching scheme. LRU and DE-UBAC show substantially larger fractions of MBS-assisted service than the RL-based schemes, indicating more frequent reliance on MBS-assisted content retrieval.

A particularly notable comparison can be made between TLPPO and GRU-DEPPO. The two schemes exhibit similar fractions of MBS-assisted service, while TLPPO achieves a slightly larger fraction of direct cache hits. However, its cooperative service is concentrated on one-hop delivery. In contrast, GRU-DEPPO also exploits two-hop and three-hop cooperative delivery, indicating that it utilizes a broader set of service opportunities available in the multi-UAV network. Despite not achieving a larger direct-hit ratio than TLPPO, GRU-DEPPO attains a substantially lower deadline-miss rate. This result indicates that cache-hit performance alone does not fully explain deadline satisfaction under heterogeneous content-specific requirements.

\begin{figure}[pos=t]
\includegraphics[width=\linewidth]{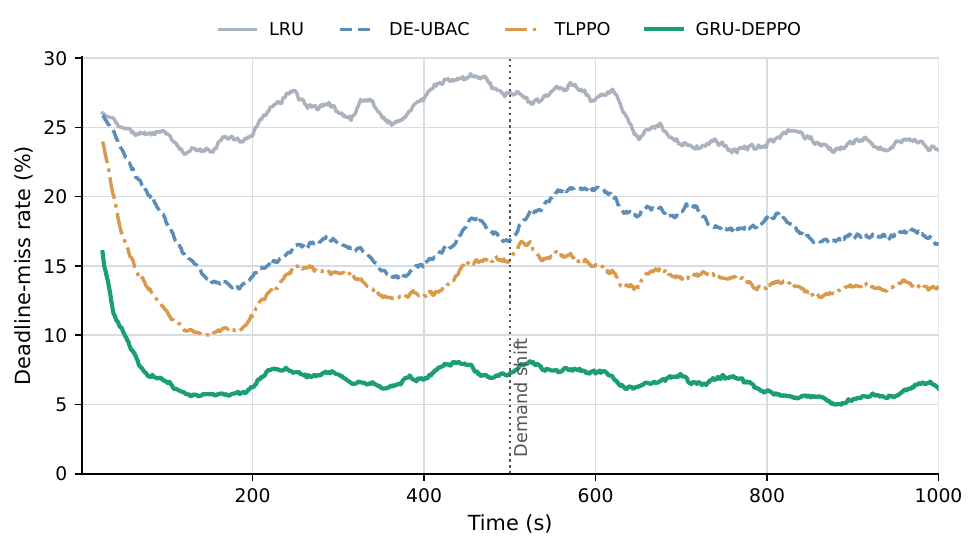}
\caption{Deadline-miss rate trend}
\label{fig:dmr_trend}
\end{figure}

We next evaluate the adaptability of the caching schemes to dynamic changes in the network environment. To facilitate a direct comparison of the adaptation behavior, the change point is deliberately fixed at the midpoint of the simulation for this experiment. Fig.~\ref{fig:dmr_trend} shows the temporal evolution of the deadline-miss rate when both the mobility context and content-demand distribution are shifted at this point. LRU and DE-UBAC exhibit relatively slow adaptation to the changed demand conditions. LRU relies on a reactive recency-based replacement rule, and DE-UBAC adapts its cache configuration using interval-level observations through predefined deadline-, cooperation-, and energy-aware rules.
However, its decisions still depend on accumulated history information, persistent-miss feedback, and fixed decision thresholds. Consequently, its cache configuration responds to an abrupt distribution shift only after the changed network conditions are sufficiently reflected in these rule-based decision signals.

In contrast, the RL-based schemes adjust their cache decisions more rapidly after the distribution shift based on the newly observed network states. Among them, GRU-DEPPO exhibits the smallest immediate DMR degradation and the fastest recovery. To quantify the immediate performance degradation, we use the average DMR during the 100~s immediately preceding the distribution shift as the pre-shift reference and compare it with the average DMR during the first 20~s after the shift. Under this criterion, the DMR increases by approximately 14.6\% for TLPPO, compared with only 7.8\% for GRU-DEPPO. The recovery time is measured as the elapsed time after the distribution shift until the DMR returns to the pre-shift reference level. GRU-DEPPO returns to this level after approximately 38~s, while TLPPO requires approximately 108~s. These results indicate that the temporal representations employed by GRU-DEPPO help the policy adapt its cache decisions more effectively to abrupt changes in content-demand conditions.

\begin{figure*}[pos=t]
\centering
\begin{subfigure}[t]{0.24\textwidth}
\centering
\includegraphics[width=\linewidth]{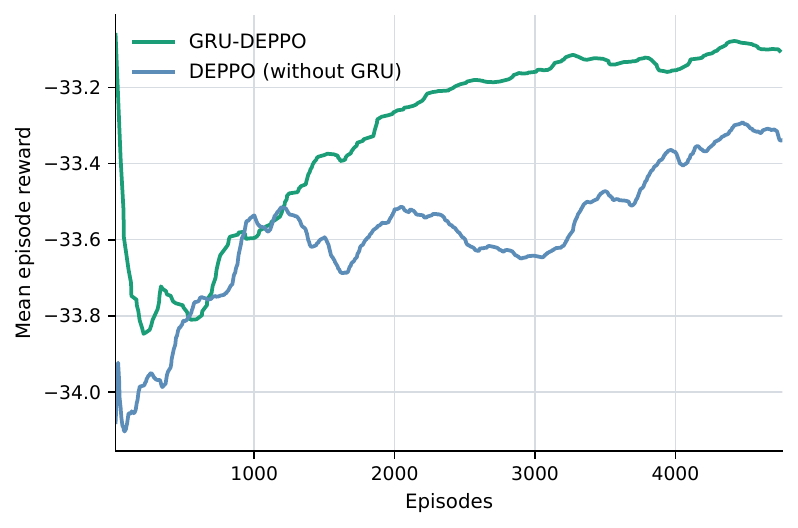}
\caption{Episode reward}
\label{fig:ablation_episode_reward}
\end{subfigure}
\hfill
\begin{subfigure}[t]{0.24\textwidth}
\centering
\includegraphics[width=\linewidth]{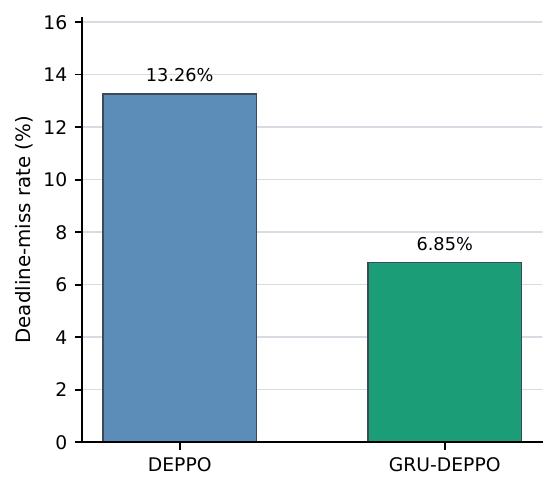}
\caption{Deadline-miss rate}
\label{fig:ablation_deadline_miss_rate}
\end{subfigure}
\hfill
\begin{subfigure}[t]{0.24\textwidth}
\centering
\includegraphics[width=\linewidth]{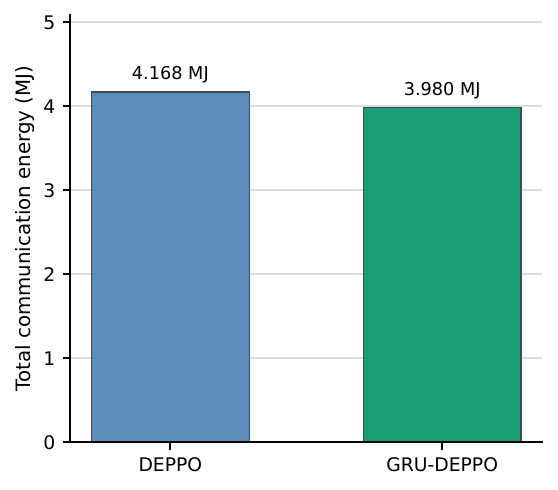}
\caption{Total communication energy}
\label{fig:ablation_total_energy}
\end{subfigure}
\hfill
\begin{subfigure}[t]{0.24\textwidth}
\centering
\includegraphics[width=\linewidth]{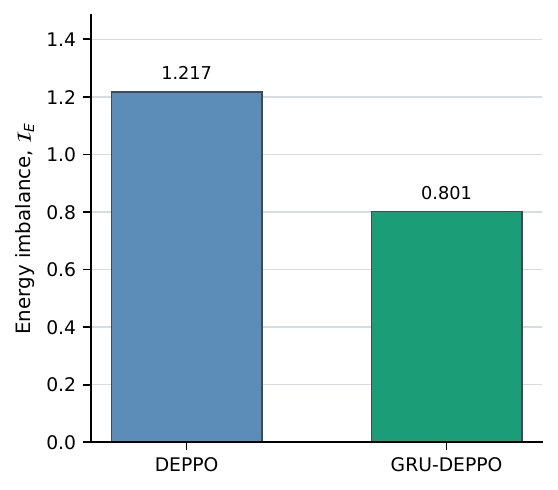}
\caption{Inter-UAV energy imbalance}
\label{fig:ablation_energy_imbalance}
\end{subfigure}
\caption{Ablation study of the GRU-based encoding}
\label{fig:gru_ablation}
\end{figure*}

Finally, we conduct an ablation study by comparing GRU-DEPPO with DEPPO, which removes the GRU-based temporal representation and directly provides the corresponding raw observations to the PPO policy. Fig.~\ref{fig:ablation_episode_reward} shows the training curves of GRU-DEPPO and DEPPO. From approximately 1,200 episodes onward, GRU-DEPPO consistently maintains a higher mean episode reward, whereas DEPPO exhibits several periods of stagnation and performance degradation during training. GRU-DEPPO also eventually reaches a higher reward level. These results indicate that the GRU-based temporal representation helps the PPO policy more effectively exploit heterogeneous and time-varying network observations, thereby facilitating the learning of a better-performing caching policy. Figs.~\ref{fig:ablation_deadline_miss_rate},\ref{fig:ablation_total_energy}, and \ref{fig:ablation_energy_imbalance} compare GRU-DEPPO and DEPPO in terms of deadline-miss rate, total communication energy consumption, and inter-UAV energy imbalance, respectively. DEPPO achieves a slightly better deadline-miss rate and lower inter-UAV energy imbalance than TLPPO, while exhibiting higher total communication energy consumption. However, clear performance gaps remain between DEPPO and GRU-DEPPO across the considered objectives.

Although DEPPO employs the same PPO-based decision framework and jointly considers deadline satisfaction, communication energy, and inter-UAV energy imbalance, it directly relies on raw network observations without GRU-based temporal encoding. As a result, the policy is less effective at extracting temporal patterns from spatially heterogeneous and time-varying network conditions. In contrast, the GRU-based representation provides a compact temporal summary of recent network observations, enabling PPO to learn more effective cache decisions. These results demonstrate the contribution of the GRU-based temporal representation to the overall performance of GRU-DEPPO.

\section{Conclusion}
\label{sec:conclusion}
In this paper, we investigated deadline- and energy-aware cache placement
and replacement in dynamic multi-UAV edge networks with spatially
heterogeneous and time-varying content demand. We first developed
DE-UBAC as an interpretable hierarchical rule-based realization of the
formulated caching objectives and used its limitations under dynamically
coupled network conditions to motivate the proposed GRU-DEPPO framework.
GRU-DEPPO employs local and global GRU-based temporal representations to
capture the evolution of UAV-local and network-wide conditions and
provides the resulting representations to a centralized PPO policy.
Simulation results showed that GRU-DEPPO achieves the lowest
deadline-miss rate and total communication energy consumption among the
considered schemes while substantially reducing inter-UAV energy
imbalance.

As future work, we plan to extend the proposed framework by jointly optimizing UAV positioning and content caching. While the present study focuses on cache placement and replacement under spatially heterogeneous and time-varying network conditions, adapting UAV positions may provide additional opportunities to improve connectivity, content accessibility, and communication-energy efficiency. 

\clearpage
\appendix
\section{Training Parameters}
\begin{table}[t]
\centering
\caption{Training parameters of GRU-DEPPO}
\label{tab:training_parameters}
\renewcommand{\arraystretch}{1.05}
\setlength{\tabcolsep}{5pt}
\footnotesize
\begin{tabularx}{\columnwidth}{
@{} >{\centering\arraybackslash}X
>{\centering\arraybackslash}X @{}
}
\hline
Parameter & Value \\
\hline

$N_{\mathrm{ep}}$ & 4,000 \\
PPO learning rate & $1\times10^{-4}$ \\
PPO optimizer & Adam \\
$B_{\mathrm{roll}}$ & 256 \\
$B_{\mathrm{mb}}$ & 128 \\
$K_{\mathrm{PPO}}$ & 3 \\
$\gamma$ & 0.99 \\
$\epsilon_{\mathrm{srv}}$ & 0.15 \\
$\lambda_{\mathrm{GAE}}$ & 0.95 \\
$\epsilon_{\mathrm{clip}}$ & 0.2 \\
$c_H$ & 0.01 \\
$c_V$ & 0.7 \\

Actor hidden layers & $(256,256)$ \\
Critic hidden layers & $(512,256)$ \\

$L$ & 10 \\
$d_{\mathrm{loc}}$ & 48 \\
Local GRU layers & 1 \\
Summary target dim. & 11 \\
Global GRU layers & 1 \\
Global token dimension & 64 \\
$d_{\mathrm{global}}$ & 32 \\

Pretraining policy & LFU \\
Pretraining optimizer & AdamW \\
Pretraining learning rate & $1\times10^{-3}$ \\
Pretraining batch size & 512 \\
Pretraining epochs & 30 \\
$\lambda_{\mathrm{type}}$ & 1 \\

\hline
\end{tabularx}
\end{table}

For compactness, $B_{\mathrm{mb}}$ denotes the PPO mini-batch size, $K_{\mathrm{PPO}}$ the number of PPO optimization epochs, $d_{\mathrm{loc}}$ and $d_{\mathrm{global}}$ the hidden dimensions of the local and global GRUs, respectively.
The simulator and GRU-DEPPO agent were implemented in Python using PyTorch and Stable-Baselines3/SB3-Contrib. Experiments were performed on an NVIDIA GeForce RTX 4090 GPU.

\section{GRU Input Composition}
\label{app:gru_input}

\begin{table}[t]
\centering
\caption{Composition of the local GRU input}
\label{tab:local_gru_input}
\renewcommand{\arraystretch}{1.05}
\setlength{\tabcolsep}{5pt}
\footnotesize
\begin{tabularx}{\columnwidth}{
@{} >{\centering\arraybackslash}X
>{\centering\arraybackslash}X @{}
}
\hline
Feature group & Dimension \\
\hline

UAV features & 20 \\
Cache-type composition & $|\mathcal{T}|$ \\
Request-type composition & $|\mathcal{T}|$ \\
Deadline pressure & $|\mathcal{T}|$ \\

\hline
Total & $20+3|\mathcal{T}|$ \\
\hline
\end{tabularx}
\end{table}

The local GRU input consists of UAV-level operating features and
content-type-specific statistics. The 20-dimensional UAV feature vector
summarizes the current association, cache, service, demand, and
communication-energy conditions of each UAV. In addition, the cache-type
composition, request-type composition, and deadline-pressure vectors each
contain $|\mathcal{T}|$ type-level entries. These features are derived
from the current UAV state and historical service and demand observations
available up to the current cache-decision epoch. Table~\ref{tab:local_gru_input}
summarizes the resulting input composition.

\begin{table}[t]
\centering
\caption{Composition of the network-wide context vector}
\label{tab:global_gru_input}
\renewcommand{\arraystretch}{1.05}
\setlength{\tabcolsep}{5pt}
\footnotesize
\begin{tabularx}{\columnwidth}{
@{} >{\centering\arraybackslash}X
>{\centering\arraybackslash}X @{}
}
\hline
Feature group & Dimension \\
\hline

Network activity & 3 \\
Service quality & 3 \\
Service-mode composition & 3 \\
Communication energy & 2 \\
Demand variation & 1 \\
Inter-UAV energy imbalance & 1 \\

\hline
Total & 13 \\
\hline
\end{tabularx}
\end{table}

In parallel with the local state encoding, the MBS constructs a compact
network-wide context vector using the current network state and observations
collected during the most recent cache-decision interval. The context vector
contains three network-activity features, three service-quality features,
three service-mode ratios, two communication-energy features, one
demand-variation feature, and one inter-UAV energy-imbalance feature.
It is defined as
\begin{equation}
\label{eq:global_gru_input}
\makebox[\linewidth][c]{%
$\displaystyle
\mathbf{c}_{\xi}^{\mathrm{net}}
=
\left[
\mathbf{g}_{\xi}^{\mathrm{act}};
\mathbf{g}_{\xi}^{\mathrm{qos}};
\mathbf{g}_{\xi}^{\mathrm{mode}};
\mathbf{g}_{\xi}^{\mathrm{energy}};
v_{\xi}^{\mathrm{dem}};
\mathcal{I}_{E,\xi}
\right]
\in\mathbb{R}^{13}
$%
}
\end{equation}

The network-wide context consists of aggregate network activity, service-quality, service-mode, communication-energy, demand-variation, and inter-UAV energy-imbalance information available at the centralized controller.

Before being processed by the global GRU, each network-context vector is projected onto a $d_{\mathrm{tok}}$-dimensional token space:
\begin{equation}
\label{eq:global_context_embedding}
\makebox[\linewidth][c]{%
$\displaystyle
\mathbf{e}_{\xi}^{\mathrm{net}}
=
\phi_{\mathrm{global}}
\left(
\mathbf{c}_{\xi}^{\mathrm{net}}
\right)
\in\mathbb{R}^{d_{\mathrm{tok}}}, d_{\mathrm{tok}}=64
$%
}
\end{equation}
The resulting token sequence is encoded by the global GRU according to~\eqref{eq:global_gru_representation}.

\clearpage

\bibliographystyle{model1-num-names}
\bibliography{cas-refs}

\end{document}